\documentclass{article}

\newcommand{\methodName}{CHAIN\xspace}

\usepackage[final]{neurips_2024}

\usepackage[utf8]{inputenc} %
\usepackage[T1]{fontenc}    %
\usepackage{hyperref}       %
\usepackage{url}            %
\usepackage{booktabs}       %
\usepackage{amsfonts}       %
\usepackage{nicefrac}       %
\usepackage{microtype}      %
\usepackage{xcolor}         %

\usepackage{microtype}
\usepackage{graphicx}
\usepackage{subfigure}

\usepackage{algorithm}
\usepackage{algorithmic}

\usepackage{amsmath}
\usepackage{amssymb}
\usepackage{mathtools}
\usepackage{amsthm}

\usepackage{multirow}
\usepackage{ulem}
\usepackage{tikz}

\usetikzlibrary{decorations.pathmorphing}

\usepackage[capitalize,noabbrev]{cleveref}

\usepackage{wrapfig}
\usepackage{comment}
\usepackage{xspace}

\usepackage{tcolorbox}
\tcbuselibrary{most}
\newtcolorbox{mybox}[2][]{colbacktitle=red!10!white, colback=gray!10!white,coltitle=black!70!black, title={#2},fonttitle=\bfseries,#1}

\title{Improving Deep Reinforcement Learning by Reducing the Chain Effect of Value and Policy Churn}

\author{%
  Hongyao Tang
  \\
  Mila - Qu\'{e}bec AI Institute\\
  Universit\'{e} de Montr\'{e}al\\
  \texttt{tang.hongyao@mila.quebec}
  \And
  Glen Berseth \\
  Mila - Qu\'{e}bec AI Institute \\
  Universit\'{e} de Montr\'{e}al \\
  \texttt{glen.berseth@mila.quebec}
}

\begin{document}

\maketitle

\begin{abstract}
Deep neural networks provide Reinforcement Learning (RL) powerful function approximators to address large-scale decision-making problems. However, these approximators introduce challenges due to the non-stationary nature of RL training. One source of the challenges in RL is that output predictions can \textit{churn}, leading to uncontrolled changes after each batch update for states not included in the batch. Although such a churn phenomenon exists in each step of network training, how churn occurs and impacts RL remains under-explored. In this work, we start by characterizing churn in a view of Generalized Policy Iteration with function approximation, and we discover a \textit{chain effect of churn} that leads to a cycle where the churns in value estimation and policy improvement compound and bias the learning dynamics throughout the iteration. Further, we concretize the study and focus on the learning issues caused by the chain effect in different settings, including greedy action deviation in value-based methods, trust region violation in proximal policy optimization, and dual bias of policy value in actor-critic methods. We then propose a method to reduce the chain effect across different settings, called Churn Approximated ReductIoN (\textbf{\methodName}), which can be easily plugged into most existing DRL algorithms. Our experiments demonstrate the effectiveness of our method in both reducing churn and improving learning performance across online and offline, value-based and policy-based RL settings, as well as a scaling setting.
\end{abstract}

\section{Introduction}
\label{sec:intro}

One fundamental recipe for the success of Deep Reinforcement Learning (DRL) is powerful approximation and generalization provided by deep neural networks, which augments the ability of RL with tabular or linear approximation to large state spaces.
However, on the other side of this benefit is less control over the function dynamics.
Network outputs can change indirectly to unexpected values after any random batch update for input data not included in the batch, called \textit{churn} in this paper. This change is particularly problematic for an RL agent due to its non-stationary nature, which can exacerbate instability, suboptimality, and even collapse.
Therefore, it is important to 
understand and control these undesired dynamics to address learning issues and improve performance.

Consistent efforts have been devoted by the RL community to gain a better understanding of the learning dynamics from different perspectives~\citep{Achiam2019Towardscharacterizing,KumarA0CTL22DR3,LiuWTJ0W23Measuring,LyleRDKG22LearningDynamics}. 
Recently, \citet{Schaul22PolicyChurn} studied a novel churn phenomenon in the learning process of typical value-based RL algorithms like DoubleDQN~\citep{HasseltGS16DoubleDQN}.
The phenomenon reveals that the greedy actions of about $10\%$ of states in the replay buffer change after a single regular batch update. Such a dramatic churn can persist throughout the learning process of DoubleDQN, causing instabilities.

In this paper, we aim to take a step further to understand how churn occurs and influences learning in different DRL settings beyond value-based RL, as well as to propose a method to control churn.
We start by formally characterizing churn in view of Generalized Policy Iteration (GPI) with function approximation to best cover most DRL settings. 
The impact of churn is two-fold in this view:
the churn in policy improvement (called the \textit{policy churn}) changes policy outputs on states that are not directly updated, while the churn in value estimation (called the \textit{value churn}) also changes the action-value landscape, thus altering greedy action and action gradient.
We then discover a \textit{chain effect of churn} that exhibits a cycle where the two types of churn compound and bias the learning dynamics throughout the iteration.
Further, we move on from the general analysis to concrete DRL settings. We focus on the learning issues caused by the chain effect
including greedy action deviation in value-based methods, trust region violation in proximal policy optimization~\citep{SchulmanWDRK17PPO} and dual bias of policy value in actor-critic methods.
The connection between the chain effect of churn and the issues necessitates an explicit control of the churn.

To this end, we propose a method called Churn Approximated ReductIoN (\textbf{\methodName}) to reduce the chain effect of churn across different DRL settings.
The main idea of \methodName is to reduce the undesirable changes to the outputs of the policy and value networks for states (and actions) outside of the current batch of data for regular DRL training.
This reduction is achieved by minimizing the change in target values for a separate batch of data when optimizing the original policy or value learning objective.
\methodName is easy to implement and plug in most existing DRL algorithms with only a few lines of code\footnote{\url{https://github.com/bluecontra/CHAIN}}.
In our experiments, we evaluate the efficacy of \methodName in a range of environments, including
MinAtar~\citep{Young2019MinAtar}, OpenAI MuJoCo~\citep{Brockman2016Gym}, DeepMind Control Suite~\citep{Tassa2018DMC} and D4RL~\citep{Fu2020D4RL}.
The results show that our method can effectively reduce churn and mitigate the learning issues, thus improving sample efficiency or final performance across online and offline, value-based and policy-based DRL settings.
Moreover, our results also show that our method helps to scale DRL agents up and achieves significantly better learning performance when using wider or deeper networks.

The main contributions of this work are summarized as follows:
(1) We study how churn occurs and influences learning from the perspective of GPI with function approximation and present the chain effect of churn. 
(2) We show how churn results in three learning issues in typical DRL settings.
(3) We propose a simple and general method and demonstrate its effectiveness in reducing churn and improving learning performance across various DRL settings and environments.

\section{Prior Work}
\label{sec:background}

In the past decade, a significant effort has been made to understand the learning issues of DRL agents and propose improvements that make DRL more stable and effective.
The early stage of this effort studied bias control for value approximation with deep neural networks introducing many improvements after DQN~\citep{MnihKSRVBGRFOPB15DQN} and DDPG~\citep{Lillicrap2015DDPG} to address overestimation or underestimation for value-based methods~\citep{HasseltGS16DoubleDQN,BellemareDMC51,HesselMHSODHPAS18RAINBOW} and deep AC methods~\citep{Fujimoto2018TD3,HaarnojaZAL18SAC,LanPFW20maxminq,KuznetsovSGV20tqc,ChenWZR21redq} respectively.
Additional works dig deeper to diagnose the learning issues regarding instability and generalization, related to the Deadly Triad in DRL~\citep{Hasselt18DRLDeadly,Achiam2019Towardscharacterizing}, stabilizing effect of target network~\citep{ZhangYW21Break,Chen2022Targetnetwork,Alexandre2022bridging}, difficulty of experience replay~\citep{SchaulQAS15PER,Kumar0L20Discor,OstrovskiCD21Passive}, over-generalization~\citep{GhiassianRLW20ImprovingbyBreaking,PanBW21FuzzyTiling,YangAA22OvercomingSpectral}, representations in DRL~\citep{Zhang21Learning,LiTZHLWMW22HyAR,tang2022PeVFA}, delusional bias~\citep{LuSB18Nondelusional,SuOLSB20Conqur}, off-policy correction~\citep{Nachum19AlgaeDice,ZhangD0S20GenDICE,Lee21OptiDICE}, interference~\citep{CobbeHKS21PPG,RaileanuF21Decouplingvp,BengioPP20Interference} and architecture choices~\citep{OtaOJMN20Canincreasing}.

One notable thread is to understand the learning dynamics of DRL agents with a focus on the non-stationary nature of RL.
A prominent phenomenon of representation ability loss is studied in~\citep{DabneyBRDQBS21VIP,IglFLBW21Transient,KumarAGL21ImplicitUnder,KumarA0CTL22DR3,Ma2023Rein}, which reveals how representations become less useful in later stages of learning, leading to myopic convergence.
Further, empirical studies in~\citep{NikishinSDBC22PrimacyBias,DOroSNBBC23Breaking,Sokar2302Dormant,Nauman2024Overestimation} demonstrate that the loss of approximation ability becomes severe and leads to collapse when high replay-ratios are adopted for better sample efficiency, while network resets and normalization methods can be simple and effective remedies.
This is further identified as plasticity loss in DRL~\citep{LyleRD22Understanding,Abbas2023Lossofplas,Dohare23Maintaining,Lyle2024Disentangling,Xu24Drm}.

Recently, it has been found that there is a dramatic change in the policy distribution where a large portion of the greedy actions change after each batch update, called \textit{policy churn}~\citet{Schaul22PolicyChurn}. 
Although intuitively related to generalization~\citep{BengioPP20Interference} and interference~\citep{LiuWTJ0W23Measuring}, it presents a lack of understanding of churn's effect on the learning behaviors of DRL agents regarding stability, convergence, exploration, etc.
\citet{Kapturowski0JRH23MEME} takes the inspiration and proposes a method to robustify the agent's behavior by 
adding an additional policy head to the value network that fits the $\epsilon$-greedy policy via policy distillation.
In this work, we further the study of churn in a more general formal framework, where churn occurs in both value and policy learning. In particular, we focus on the dynamics of churn during the learning process and how it incurs issues in different DRL algorithms and propose a practical method to reduce churn and improve learning performance.

\section{Preliminaries}
\label{subsec:preliminaries}

Reinforcement Learning (RL) is formulated within the framework of a Markov Decision Process (MDP) $\left< \mathcal{S}, \mathcal{A}, \mathcal{P}, \mathcal{R},  \gamma, \rho_0, T \right>$,
defined with the state set $\mathcal{S}$, the action set $\mathcal{A}$, the transition function $\mathcal{P}: \mathcal{S} \times \mathcal{A} \rightarrow P(\mathcal{S})$, the reward function $\mathcal{R}: \mathcal{S} \times \mathcal{A} \rightarrow \mathbb{R}$, the discounted factor $\gamma \in [0,1)$, the initial state distribution $\rho_0$ and the horizon $T$.
The agent interacts with the MDP by performing actions from its policy $a_t \sim \pi(s_t) $ that defines the mapping from states to actions or action distributions.
The objective of an RL agent is to optimize its policy to maximize the expected discounted cumulative reward
$J(\pi) = \mathbb{E}_{\pi} [\sum_{t=0}^{T}\gamma^{t} r_t ]$,
where $s_{0} \sim \rho_{0}\left(s_{0}\right)$, $s_{t+1} \sim \mathcal{P}\left(s_{t+1} \mid s_{t}, a_{t}\right)$ and $r_t = \mathcal{R}\left(s_{t},a_{t}\right)$.
The state-action value function $q^{\pi}$ defines 
the expected cumulative discounted reward for all $s,a \in \mathcal{S} \times \mathcal{A}$ and the policy $\pi$,
i.e., $q^{\pi}(s, a)=\mathbb{E}_{\pi} \big[\sum_{t=0}^{T} \gamma^{t} r_{t} \mid s_{0}=s, a_{0}=a \big]$.

Policy and value functions are approximated with deep neural networks to cope with large and continuous state-action space.
Conventionally, $q^{\pi}$ can be approximated by $Q_{\theta}$ with parameters $\theta$
typically through minimizing Temporal Difference (TD) loss \citep{Sutton1988ReinforcementLA}, 
i.e., 
$L(\theta)= \mathbb{E}_{s,a \sim D} \ \delta_{\theta}(s,a)^2$ where $D$ is a replay buffer and $\delta_{\theta}(s,a)$ is a type of TD error.
A parameterized policy $\pi_{\phi}$ with parameters $\phi$ can be updated by taking the gradient of the objective, 
i.e., $\phi^{\prime} \leftarrow \phi + \alpha \nabla_{\phi} J(\pi_{\phi})$ with a step size $\alpha$.
Value-based methods like Deep $Q$-Network (DQN)~\citep{MnihKSRVBGRFOPB15DQN} trains a $Q$-network $Q_{\theta}$ by minimizing $L(\theta)$ where $\delta(s,a) = Q_{\theta}(s,a) - (r + \gamma \max_{a^{\prime}}Q_{\theta^{-}}(s^{\prime},a^{\prime}))$ and $\theta^{-}$ denotes the target network.
For policy-based methods, TD3~\citep{Fujimoto2018TD3} is often used to update a deterministic policy with Deterministic Policy Gradient (DPG) theorem \citep{Silver2014DPG}:
$\nabla_{\phi} J(\pi_{\phi}) = \mathbb{E}_{s \sim D} \left[ \nabla_{\phi} \pi_{\phi}(s) \nabla_{a} Q_{\theta}(s,a)|_{a=\pi_{\phi}(s)}\right]$;
Soft Actor-Critic (SAC)~\citep{HaarnojaZAL18SAC} learns a stochastic policy with the gradient:
$\nabla_{\phi} \hat{J}(\pi_{\phi}) = \mathbb{E}_{s \sim D} \big[ \nabla_{\phi} \log \pi_{\phi}(a|s) + (\nabla_{a} \log \pi_{\phi}(a|s)) - \nabla_{a} Q_{\theta}(s,a)) \nabla_{\phi} f_{\phi}(\epsilon ;s)) |_{a=f_{\phi}(\epsilon ; s)}\big]$, with noise $\epsilon$ and implicit function $f_{\phi}$ for re-parameterization.

\section{A Chain Effect of Value and Policy Churn}
\label{sec:churn}

In this section, we present a formal study on value and policy churn and their impact on learning.
We first introduce an intuitive overview of how churn is involved in DRL
(\Cref{subsec:gpi_under_churn}).
Then, we propose the definitions of the value and policy churn (\Cref{subsec:def_of_churn}),
followed by a chain effect that reveals how the churns interplay and bias parameter update (\Cref{subsec:chain_effect}).

\subsection{Generalized Policy Iteration under Churn}
\label{subsec:gpi_under_churn}

\begin{wrapfigure}{r}{0cm}
\begin{minipage}[t]{.4\linewidth}
\centering
\vspace{-0.8cm}
\hspace{-0.1cm}
\scalebox{0.55}{
\begin{tikzpicture}
\node at (3,3.8) {\Large Evaluation};
\node at (3,3) {\Large $Q_{\theta} \leadsto q^{\pi_{\phi}}$};
\node at (3,-1.2) {\Large Improvement};
\node at (3,-0.4) {\Large $\pi_{\phi} \leadsto \text{greedy}(Q_{\theta})$};
\node[color=red] at (6.8,1.5) {\large Value};
\node[color=red] at (6.8,1.0) {\large Churn};
\node[color=red] at (-0.8,1.5) {\large Policy};
\node[color=red] at (-0.8,1.0) {\large Churn};
\node[rectangle,
rounded corners =3pt,
minimum width =30pt,
minimum height =30pt] (1) at(0,2.5){};
\node at (0,2.5) {\Large ${\pi}_{\phi}$};
\node[rectangle,
rounded corners =3pt,
minimum width =30pt,
minimum height =30pt] (8) at(0.3,2.5){};
\node[rectangle,
rounded corners =3pt,
minimum width =30pt,
minimum height =30pt] (2) at(6,2.5){};
\node[rectangle,
rounded corners =3pt,
minimum width =30pt,
minimum height =30pt] (3) at(6.3,2.5){};
\node at (6,2.5) {\Large $\bar Q_{\theta}$};
\node[rectangle,
rounded corners =3pt,
minimum width =30pt ,
minimum height =30pt] (4) at(5.7,0){};
\node[rectangle,
rounded corners =3pt,
minimum width =30pt ,
minimum height =30pt] (5) at(6,0){};
\node at (6,0) {\Large $Q_{\theta}$};
\node[rectangle,
rounded corners =3pt,
minimum width =30pt ,
minimum height =30pt] (6) at(0,0){};
\node[rectangle,
rounded corners =3pt,
minimum width =30pt ,
minimum height =30pt] (7) at(-0.3,0){};
\node at (0,0) {\Large $\bar \pi_{\phi}$};
\draw[line width=1.5pt, ->, scale=2] (1) to[bend left=60] (2);
\draw[line width=1pt, ->, scale=2, color=red] (3) to [out=240, in=60, looseness=1.5] (4);
\draw[line width=1.5pt, ->, scale=2] (5) to[bend left=60] (6);
\draw[line width=1pt, ->, scale=2, color=red] (7) to [out=60, in=240, looseness=1.5] (8);
\end{tikzpicture}
}
\vspace{-0.4cm}
\caption{Generalized Policy Iteration (GPI) under the value and policy churn.
}
\label{figure:gpi_with_churn}
\vspace{-0.6cm}
\end{minipage}
\end{wrapfigure}
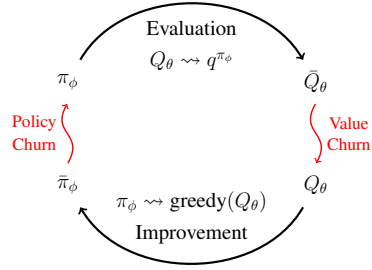

Generalized Policy Iteration (GPI)~\citep{Sutton1988ReinforcementLA} is widely used to refer to the general principle of learning in an Evaluation-Improvement iteration manner,
which applies to almost all RL methods.
In the context of DRL, 
i.e., with network representation and mini-batch training,
the value and policy networks' outputs can have unexpected changes, i.e., the \textit{churn}, after each mini-batch training for the states not included in the batch.
Such churns are neglected in most DRL methods, let alone their influence on the practical learning process.
In Figure~\ref{figure:gpi_with_churn}, we extend the classic GPI diagram by considering churn to show how it is involved in the learning process intuitively.

In the evaluation process, the parameterized $Q$-network $Q_{\theta}$ approximates the value of the current policy via repeated mini-batch training.
Under the impact of churn, the $Q$-network is not likely to have output predictions the same as what was updated with explicit mini-batch training.
For example, let's imagine we have a \textit{virtual} network $\bar Q_{\theta}$ that only accepts the changes for the states updated by mini-batch training directly and remains unchanged for the others.

Thus, the \textit{value churn}, denoted by $\bar Q_{\theta} \leadsto Q_{\theta}$, is an implicit process that alters the virtual network $\bar Q_{\theta}$ to the approximation $Q_{\theta}$ we obtained in practice.
Similarly, the \textit{policy churn} $\bar \pi_{\phi} \leadsto {\pi}_{\phi}$ occurs in the improvement process.
As illustrated in Figure~\ref{figure:gpi_with_churn}, the value churn and the policy churn are interwoven in the Evaluation-Improvement process. Usually, we can assume that the churns make non-negligible changes, i.e., $Q_{\theta} \ne \bar{Q}_{\theta}$ and $\pi_{\phi} \ne \bar{\pi}_{\phi}$.
Therefore, we delve into the cause of churn, its impact on learning, and possible remedies to mitigate its negative impact in the following sections.

\subsection{Definition of Value and Policy Churn}
\label{subsec:def_of_churn}

A deep neural network can have the form $f_{\theta}: X \rightarrow Y$ with parameters $\theta$. The network is optimized for a set of input data $B_{\text{train}}=\{x_i\}$ with a loss function, leading to a parameter update of $\theta \rightarrow \theta^{\prime}$. 
Given a reference set of input data $B_{\text{ref}} = \{\bar x_i\}$ (where $B_{\text{ref}} \cap B_{\text{train}} = \emptyset$) and a metric $d$ for the output space, the churn is formally defined as:
\begin{equation*}
    \mathcal{C}_{f}(\theta,\theta^{\prime}, B_{\text{ref}}) = \frac{1}{|B_{\text{ref}}|}\sum_{\bar x \in B_{\text{ref}}} d(f_{\theta^{\prime}}(\bar x), f_{\theta}(\bar x)).
\end{equation*}
Arguably, churn is an innate property of neural networks, and it is closely related to problems like interference~\citep{Liu20Towards,LiuWTJ0W23Measuring} and catastrophic forgetting~\citep{LanPLM23Memory} in different contexts. 

In this paper, 
we focus on the churn in $Q$-value network $Q_{\theta}$ and policy network $\pi_{\phi}$.
We then obtain the definitions of the $Q$-value churn ($\mathcal{C}_{Q}$, w.r.t. $\theta \rightarrow$ \textcolor{blue}{$\theta^{\prime}$}) and the policy churn ($\mathcal{C}_{\pi}$, w.r.t. $\phi \rightarrow$ \textcolor{blue}{$\phi^{\prime}$}, using a deterministic policy for demonstration) for an arbitrary state-action pair $\bar s, \bar a \in B_{\text{ref}}$ as follows:
\begin{equation}
\begin{aligned}
    c_{Q}(\theta,\textcolor{blue}{\theta^{\prime}}, \bar s, \bar a) =  Q_{\textcolor{blue}{\theta^{\prime}}}(\bar s, \bar a) - Q_{\theta}(\bar s, \bar a) & , \ c_{\pi}(\phi,\textcolor{blue}{\phi^{\prime}}, \bar s) = \pi_{\textcolor{blue}{\phi^{\prime}}}(\bar s) - \pi_{\phi}(\bar s) .%
\end{aligned}
\end{equation}
Then, the definitions regarding $B_{\text{ref}}$ can be generalized to the batch setting by aggregating data in $B_{\text{ref}}$: $\mathcal{C}_{Q}(\theta,\textcolor{blue}{\theta^{\prime}}, B_{\text{ref}})  = \frac{1}{|B_{\text{ref}}|}\sum_{\bar s, \bar a \in B_{\text{ref}}} |c_Q(\theta,\textcolor{blue}{\theta^{\prime}}, \bar s, \bar a)|$, $
\mathcal{C}_{\pi}(\phi,\textcolor{blue}{\phi^{\prime}}, B_{\text{ref}}) = \frac{1}{|B_{\text{ref}}|}\sum_{\bar s \in B_{\text{ref}}} |c_{\pi}(\phi,\textcolor{blue}{\phi^{\prime}}, \bar s)|$.
Without loss of generality, we carry out our analysis mainly regarding $\bar s, \bar a$ for clarity in the following.

\vspace{-0.2cm}
\paragraph{(U.I) How the churns $\mathcal{C}_{Q}, \mathcal{C}_{\pi}$ are caused by parameter updates} 
First, we look into the relationship between the $Q$-value churn $\mathcal{C}_{Q}$, the policy churn $\mathcal{C}_{\pi}$ and the network parameter updates $\Delta_{\theta} = \theta^{\prime} - \theta, \Delta_{\phi} =\phi^{\prime} - \phi$.
For $\Delta_{\theta}, \Delta_{\phi}$,
we use typical TD learning and DPG for demonstration:
$\Delta_{\theta} = \frac{\alpha}{|B_{\text{train}}|} \sum_{s,a \in B_{\text{train}}} \nabla_{\theta}Q_{\theta}(s,a) \delta_{\theta}(s,a)$, and $\Delta_{\phi} = \frac{\alpha}{|B_{\text{train}}|} \sum_{s \in B_{\text{train}}} \nabla_{\phi}\pi_{\phi}(s) \nabla_{a}Q_{\theta}(s,a)|_{a=\pi_{\phi}(s)}$.

Now we characterize $\mathcal{C}_{Q}$ and $\mathcal{C}_{\pi}$ as functions of $\Delta_{\theta}, \Delta_{\phi}$ with the help of Neural Tangent Kernel (NTK)~\citep{Achiam2019Towardscharacterizing}.
For clarity, we use $B_{\text{train}} = \{s, a\}$ and $B_{\text{ref}} = \{\bar s, \bar a\}$ and abbreviate $B_{\text{train}}, B_{\text{ref}}$ and step size $\alpha$ when context is clear. Concretely,
\begin{equation}
\begin{aligned}
    c_{Q}(\theta, \theta^{\prime}) 
    & = \nabla_{\theta}Q_{\theta}(\bar s, \bar a)^{\top}\Delta_{\theta} + O(\|\Delta_{\theta}\|^2) \approx \underbrace{\nabla_{\theta}Q_{\theta}(\bar s, \bar a)^{\top}\nabla_{\theta}Q_{\theta}(s,a) }_{k_{\theta}(\bar s, \bar a, s, a)} \delta_{\theta}(s,a) \\
    c_{\pi}(\phi, \phi^{\prime}) 
    & = \nabla_{\phi}\pi_{\phi}(\bar s)^{\top}\Delta_{\phi} + O(\|\Delta_{\phi}\|^2) \approx \underbrace{\nabla_{\phi}\pi_{\phi}(\bar s)^{\top} \nabla_{\phi}\pi_{\phi}(s) }_{k_{\phi}(\bar s, s)} \nabla_{a}Q_{\theta}(s,a)|_{a=\pi_{\phi}(s)}
\end{aligned}
\label{eq:churn_caused_by_parameter_update}
\end{equation}
Eq.~\ref{eq:churn_caused_by_parameter_update} shows that the value and policy churn are mainly determined by the kernels of the $Q$-network $k_{\theta}$ and the policy network $k_{\phi}$, along with the TD error and the action gradient.
This indicates that churn is determined by both the network's property itself and the learning we performed with the network.

\subsection{From Single-step Interplay to The Chain Effect of Churn}
\label{subsec:chain_effect}

In addition to the first piece of understanding \textbf{(U.I)} that presents how parameter updates cause the churns, we discuss how the churns affect parameter updates backward with two more pieces of understanding \textbf{(U.II)} and \textbf{(U.III)}, finally shedding light on a chain effect of churn.

\vspace{-0.2cm}
\paragraph{(U.II) How $\mathcal{C}_{Q}, \mathcal{C}_{\pi}$ deviates action gradient and policy value} 

First, we introduce two types of deviation derived from the value and policy churn:
(1) Action Gradient Deviation ($\mathcal{D}^{Q}_{\nabla_{a}}$), the change of action gradient regarding the $Q$-network for states and actions that are affected by the $Q$-value churn $\mathcal{C}_{Q}$;
(2) Policy Value Deviation ($\mathcal{D}^{\pi}_{Q}$), the change of $Q$-value due to the action change for states that are affected by policy churn $\mathcal{C}_{\pi}$.
Formally,
the two types of deviation are:
\begin{equation}
\begin{aligned}
    d^{Q}_{\nabla_{a}}(\theta,\textcolor{blue}{\theta^{\prime}}, \bar s) = & \ \nabla_{\bar a} Q_{\textcolor{blue}{\theta^{\prime}}}(\bar s,\bar a)|_{\bar a=\pi(\bar s)}
    - \nabla_{\bar a} Q_{\theta}(\bar s,\bar a)|_{\bar a=\pi(\bar s)}. \\
    d^{\pi}_{Q}(\phi,\textcolor{blue}{\phi^{\prime}}, \bar s) = & \ Q(\bar s,\pi_{\textcolor{blue}{\phi^{\prime}}}(\bar s))- Q(\bar s,\pi_{\phi}(\bar s)). 
\end{aligned}
\end{equation}
One thing to note is the two types of deviation show the interplay between the value and policy churn, as the value churn derives the deviation in policy ($c_{Q} \xrightarrow{\text{derive}} d^{Q}_{\nabla_{a}}$) and the policy churn derives the deviation in value ($c_{\pi} \xrightarrow{\text{derive}} d^{\pi}_{Q}$), as denoted by the superscripts.
This interplay between the policy and value can be shown better with the expressions below (derivation details in Appendix~\ref{app:derivations}):
\begin{equation}
\begin{aligned}
    d^{Q}_{\nabla_{a}}(\theta,\theta^{\prime}) & = \nabla_{\bar a} c_{Q}(\theta, \theta^{\prime})|_{\bar a=\pi(\bar s)}, 
    \ \ d^{\pi}_{Q} (\phi, \phi^{\prime}) 
    \approx (\nabla_{\bar a} Q_{\theta}(\bar s, \bar a)|_{\bar a=\pi_{\phi}(\bar s)})^{\top} 
    c_{\pi}(\phi, \phi^{\prime})
\end{aligned}
\label{eq:churn_derivatives}
\end{equation}
Since the action gradient and policy value play key roles in parameter updates, the deviations in them naturally incur negative impacts on learning.

The discussion thus far is within a single parameter update.
Now, we discuss the implications of these single updates towards a chain of updates to shed light on the long-term effect of churn.

\vspace{-0.2cm}
\paragraph{(U.III) How parameter updates are biased by $\mathcal{C}_{Q}, \mathcal{C}_{\pi}$ and the deviations $\mathcal{D}^{Q}_{\nabla_{a}}, \mathcal{D}^{\pi}_{Q}$}

Let us consider a segment of two consecutive updates, denoted by $(\theta^{-}, \phi^{-}) \rightarrow (\theta, \phi) \rightarrow (\theta^{\prime}, \phi^{\prime})$.
The churns occurred during the \textit{last update} $(\theta^{-}, \phi^{-}) \rightarrow (\theta, \phi)$ participate in the \textit{current update} $(\theta, \phi) \rightarrow (\theta^{\prime}, \phi^{\prime})$ about to perform.
Concretely, the churns affect the following aspects:
(1) $Q$-value estimate, (2) action selection in both TD error and policy objective,
and (3) the gradient of network parameters.

From these aspects, we can deduce the difference between the parameter updates under the impact of the value and policy churn
(denoted by $\tilde{\Delta}_{\theta},\tilde{\Delta}_{\phi}$) 
and the conventional ones $\Delta_{\theta},\Delta_{\phi}$.
As a result, we can find that the value and policy churn, as well as the deviations derived, introduce biases in the parameter updates.
We provide the complete discussion and derivation in~\Cref{app:chain_effect_detail}.

The analysis on the update segment $(\theta^{-}, \phi^{-}) \rightarrow (\theta, \phi) \rightarrow (\theta^{\prime}, \phi^{\prime})$
can be forwarded, and taking the three pieces of understanding together, 
we arrive at the chain effect of churn.
\begin{mybox}[detach title,before upper={\tcbtitle\quad}]{The Chain Effect of Churn:}
\textbf{(U.I)} Parameter updates cause the value and policy churn, \textbf{(U.II)} which further leads to the deviations in the action gradient and policy value;
\textbf{(U.III)} the churns and the deviations then bias following parameter updates.
\end{mybox}

\begin{wrapfigure}{r}{0cm}
\begin{minipage}[t]{.55\linewidth}
\centering
\vspace{-0.5cm}
\hspace{-0.1cm}
\scalebox{0.65}{
\begin{tikzpicture}
\node[rectangle,
minimum width =30pt ,
minimum height =10pt] at(6.2,3.2){Parameter Update};
\node[rectangle,
minimum width =30pt ,
minimum height =10pt] at(6.2,2.7){of $Q$ and Policy};
\node[rectangle,
minimum width =30pt ,
minimum height =10pt] at(0,-1.1){$Q$-value and Policy Churn};
\node[rectangle,
minimum width =30pt ,
minimum height =10pt] at(6,-1.1){Deviation in Action Gradient and Policy Value};
\node[rectangle,
minimum width =30pt ,
minimum height =15pt] at(0.7,1.65){Cause};
\node[rectangle,
minimum width =30pt ,
minimum height =15pt] at(2.75,-0.4){Derive};
\node[rectangle,
minimum width =30pt ,
minimum height =15pt] at(3.65,1.65){Influence};
\node[rectangle,
rounded corners =3pt,
minimum width =30pt ,
minimum height =30pt ,draw=black] (1) at(3,3){$\tilde{\Delta}_{\theta_{t}}(B_{\text{train}})$, $\tilde{\Delta}_{\phi_{t}}(B_{\text{train}})$};
\node[rectangle,
rounded corners =3pt,
minimum width =90pt ,
minimum height =40pt ,draw=red] (2) at(0,0){};
\node[rectangle,
rounded corners =3pt,
minimum width =30pt ,
minimum height =10pt] at(0,0.3){\textcolor{red}{$\mathcal{C}_{Q}(\theta_{t},\theta_{t+1}, \{\bar s, \bar a \}) $}};
\node[rectangle,
rounded corners =3pt,
minimum width =30pt ,
minimum height =10pt] at(0,-0.3){\textcolor{red}{$\mathcal{C}_{\pi}(\phi_{t},\phi_{t+1}, \{\bar s, \bar a\})$}};
\node[rectangle,
rounded corners =3pt,
minimum width =120pt ,
minimum height =40pt ,draw=red] (3) at(6,0){};
\node[rectangle,
rounded corners =3pt,
minimum width =30pt ,
minimum height =10pt] at(6,0.3){\textcolor{red}{$\mathcal{D}^{Q}_{\nabla_{a}}(\theta_{t},\theta_{t+1}, \{s, \pi_{\phi_{t}}(s)\})$}};
\node[rectangle,
rounded corners =3pt,
minimum width =30pt ,
minimum height =10pt] at(6,-0.3){\textcolor{red}{$\mathcal{D}^{\pi}_{Q}(\phi_{t},\phi_{t+1}, \{s^{\prime}, \pi_{\phi_{t}}(s^{\prime})\})$}};
\draw[line width=1.5pt, ->, scale=4] (1) --(2);
\draw[line width=1.5pt, ->, scale=4] (2) --(3);
\draw[line width=1.5pt, ->, scale=4] (3) --(1);
\draw[line width=1.5pt, ->, scale=4] (2) to[bend right=40] (1);
\end{tikzpicture}
}
\vspace{-0.3cm}
\caption{Illustration of the logical cycle of the \textit{chain effect} of the value and policy churn.}
\label{figure:vicious_cycle}
\vspace{-0.6cm}
\end{minipage}
\end{wrapfigure}
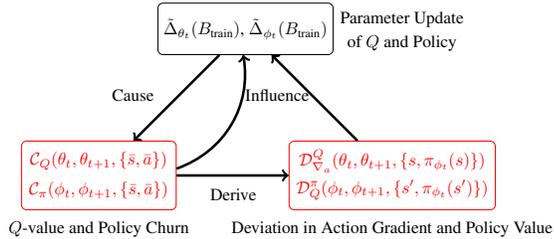
As the cycle illustrated in Figure~\ref{figure:vicious_cycle},
the value and policy churn and the parameter update bias \textit{accumulate} and can \textit{amplify each other} throughout the learning process.
Intuitively, the parameter update chain could derail and fluctuate under the accumulating churns and biases,
thus preventing stable and effective learning.
We concretize our study on the consequences in the next section.

\section{Reducing Value and Policy Churn in Deep RL}
\label{sec:churn_reduction_method} 

In this section, we show concrete learning issues caused by churn in typical DRL scenarios (\Cref{subsec:consequences_of_churn}), followed by a simple plug-in method to reduce churn and address the issues (\Cref{subsec:chain_method}).

\subsection{Consequences of the Chain Effect of Churn in Different DRL Scenarios}
\label{subsec:consequences_of_churn}

Since churn is involved in most DRL methods, as illustrated by Figure~\ref{figure:gpi_with_churn}, 
our study focuses on several typical DRL scenarios below.

\vspace{-0.2cm}
\paragraph{Greedy action deviation in value-based methods}

Value-based methods like DQN 
train a $Q$-network $Q_{\theta}$
and compute the policy by choosing the greedy action of $Q_{\theta}$. A consequence of computing the action greedily is that changes in the values will directly cause changes in the action distribution~\citep{Schaul22PolicyChurn}.
Similarly to Eq.~\ref{eq:churn_derivatives}, this deviation can be formalized as: $\mathcal{D}^{Q}_{a^{*}}(\theta, \textcolor{blue}{\theta^{\prime}}, B_{\text{ref}}) = \frac{1}{|B_{\text{ref}}|}\sum_{\bar s \in B_{\text{ref}}} \mathbb{I}_{ \mathcal{A} \backslash \{ \arg\max_a Q_{\theta}(\bar s,a) \} } \big(\arg\max_{\bar a} Q_{\textcolor{blue}{\theta^{\prime}}}(\bar s,\bar a) \big)$.
We suspect that this deviation introduces instability and hinders learning, and we focus on whether reducing churn can improve the performance of value-based methods.

\vspace{-0.2cm}
\paragraph{Trust region violation in policy gradient methods} Trust region plays a critical role in many policy gradient methods for reliable and efficient policy updates.
Proximal Policy Optimization (PPO)~\citep{SchulmanWDRK17PPO} uses a clipping mechanism as a simple but effective surrogate of the trust region for TRPO~\citep{SchulmanLAJM15TRPO}: \texttt{Clip}$\big(r(\phi_{\text{old}},\phi), 1 - \epsilon, 1+\epsilon \big)$ and
$r(\phi_{\text{old}},\phi) = \frac{\pi_{\phi}(a|s)}{\pi_{\phi_{\text{old}}}(a|s)}$.
With respect to policy churn, even though the PPO policy conforms to the trust region for the states in the current training batch, it could silently violate the trust region for other states, including previously updated ones.
Consider the policy update $\phi \rightarrow \phi^{\prime}$, it is highly likely to have $r(\phi,\textcolor{blue}{\phi^{\prime}}) = \frac{\pi_{\textcolor{blue}{\phi^{\prime}}}(\bar a| \bar s)}{\pi_{\phi}(\bar a| \bar s)} \ne 1$
and thus $r(\phi_{\text{old}},\textcolor{blue}{\phi^{\prime}}) = \frac{\pi_{\textcolor{blue}{\phi^{\prime}}}(\bar a|\bar s)}{\pi_{\phi_{\text{old}}}(\bar a|\bar s)} = r(\phi_{\text{old}}, \phi) r(\phi, \textcolor{blue}{\phi^{\prime}}) \ne r(\phi_{\text{old}}, \phi)$.
Since we have no information about $r(\phi,\textcolor{blue}{\phi^{\prime}})$, there is no guarantee for the trust region $1 - \epsilon \le r(\phi_{\text{old}}, \textcolor{blue}{\phi^{\prime}}) \le 1 + \epsilon$ to be respected after churn.
Intuitively, this implicit violation is hazardous and detrimental to learning.

\vspace{-0.2cm}
\paragraph{Dual bias of policy value in Actor-Critic methods}

Deep AC methods interleave the training between the actor-network and the critic-network.
Unlike the two scenarios above, where either the value churn or the policy churn raises a learning stability issue, we present the dual bias of policy value that stems from the bilateral effect of churn.
The dual bias exists in the policy value as $Q_{\textcolor{blue}{\theta^{\prime}}}(\bar s,\pi_{\textcolor{blue}{\phi^{\prime}}}(\bar s)) \ne Q_{\theta}(\bar s,\pi_{\phi}(\bar s))$.
In the context of AC methods, the policy value is used for the target computation of the critic $r_{t} + \gamma Q_{\textcolor{blue}{\theta^{\prime}}}(s_{t+1},\pi_{\textcolor{blue}{\phi^{\prime}}}(s_{t+1}))$ and the optimization objective of the actor $\nabla_{\phi}Q_{\textcolor{blue}{\theta^{\prime}}}(s,\pi_{\textcolor{blue}{\phi^{\prime}}}(s))$.
Thus, the dual bias steers the training of the actor and the critic.

Given these negative consequences of churn, a question is raised naturally:
how can we control the level of churn to mitigate the issues without introducing complex trust regions or constraints?

\subsection{A Regularization Method for Churn Reduction}
\label{subsec:chain_method}

In this section, we propose a regularization method to reduce value and policy churn, called Churn Approximated ReductIoN (\textbf{\methodName}).
To combat the prevalence of churn's negative influence on DRL, our method should be simple to implement and easy to use with different RL methods.

Based on the definitions of the value churn ($\mathcal{C}_{Q}$) and the policy churn ($\mathcal{C}_{\pi}$) in~\Cref{subsec:def_of_churn}, 
we propose two corresponding loss functions $L_{\text{QC}}$ and $L_{\text{PC}}$ for churn reduction.
Formally, for parameterized networks $Q_{\theta_{t}},\pi_{\phi_{t}}$ at time $t$ and a reference batch $B_{\text{ref}}$ sampled from replay buffer, we have:
\begin{equation}
\small
\hspace{-0.2cm}
    L_{\text{QC}}(\theta_{t}, B_{\text{ref}}) = \frac{1}{|B_{\text{ref}}|}\sum_{\bar s, \bar a \in B_{\text{ref}}} \left( Q_{\theta_t}(\bar s, \bar a) -  Q_{\theta_{t-1}}(\bar s, \bar a)  \right)^2
\label{eq:l_qc}
\end{equation}
\begin{equation}
\small
    L_{\text{PC}}(\phi_{t}, B_{\text{ref}}) = \frac{1}{|B_{\text{ref}}|}\sum_{\bar s \in B_{\text{ref}}} d_{\pi}(\pi_{\phi_t}(\bar s),  \pi_{\phi_{t-1}}(\bar s) )
\label{eq:l_pc}
\end{equation}
where $d_{\pi}$ is a policy distance metric, and we use mean square error or KL divergence for deterministic or stochastic policies.
Ideally, the regularization should be imposed on the post-update network parameters $\theta_{t+1}, \phi_{t+1}$. Since they are not available at time $t$, we regularize $\theta_{t}, \phi_{t}$ and use $\theta_{t-1}, \phi_{t-1}$ as the targets for a convenient and effective surrogate.

By minimizing $L_{\text{QC}}$ and $L_{\text{PC}}$, we can reduce the value and policy churn and suppress the chain effect further.
This allows us to use the churn reduction regularization terms along with standard RL objectives and arrives at DRL with \methodName finally:  
\begin{equation}
    \text{minimize}_{\theta} \ L(\theta_{t}, B_{\text{train}}) + \lambda_{Q} L_{\text{QC}}(\theta_{t}, B_{\text{ref}})
\label{eq:q_objectives}
\end{equation}
\begin{equation}
\begin{aligned}
    \text{maximize}_{\phi} \ & J(\phi_t, B_{\text{train}}) - \lambda_{\pi} L_{\text{PC}}(\phi_{t}, B_{\text{ref}})
\end{aligned}
\label{eq:p_objectives}
\end{equation}
where $B_{\text{train}}, B_{\text{ref}}$ are two separate batches randomly sampled from $D$, and $\lambda_{Q}, \lambda_{\pi}$ are coefficients that control the degree of regularization.
\methodName serves as a plug-in component that can be implemented with only a few lines of code modification in most DRL methods. 
The pseudocode is omitted here, and we refer the readers to Algorithm~\ref{alg:drl_with_chain} in the Appendix.

\paragraph{Automatic adjustment of $\lambda_{Q},\lambda_{\pi}$}
To alleviate the difficulty of manually selecting the regularization coefficients, we add a simple but effective method to adjust $\lambda_{Q},\lambda_{\pi}$ adaptively during the learning process.
The key principle behind this is to keep a consistent relative scale (denoted by $\beta$) between the churn reduction regularization terms and the original DRL objectives.
More precisely, by maintaining the running means of the absolute $Q$ loss $| \bar L_{Q}|$ and the VCR term $| \bar  L_{\text{QC}} | $, $\lambda_{Q}$ is computed dynamically as $\lambda_{Q} = \beta \frac{| \bar L_{Q}| }{| \bar  L_{\text{QC}} | }$, which is similar for $\lambda_{\pi}$. 
This is inspired by our empirical observations and the recent study on addressing the reward scale difference across different domains~\citep{Dreamerv3}.

Another thing worth noting is that \methodName helps to mitigate the loss of plasticity via churn reduction.
This connection can be established by referring to the NTK expressions in Eq.~\ref{eq:churn_caused_by_parameter_update}: reducing churn encourages $k_{\theta}, k_{\phi}$ to 0 and thus prevents the empirical NTK matrix from being low-rank, which is shown to be a consistent indicator of plasticity loss~\citep{Lyle2024Disentangling}.

\section{Experiments}
\label{sec:exps}

In the experiments, we aim to answer the following questions:
(1) How large is the value and policy churn in practice, and can our method effectively reduce churn?
(2) Does our method's reduction of churn address learning issues and improve performance in terms of efficiency and episode return?
(3) Does CHAIN also improve the scaling abilities of deep RL?

We organize our experiments into the four subsections below that correspond to the three DRL scenarios discussed in~\Cref{subsec:consequences_of_churn} as well as a DRL scaling setting.
Our experiments include 20 online RL tasks from MinAtar, MuJoCo, DMC, and 8 offline RL datasets from D4RL, as well as 6 popular algorithms, i.e., DoubleDQN, PPO, TD3, SAC, IQL, AWAC.
We provide the experimental details in Appendix~\ref{app:exp_details} and more results in Appendix~\ref{app:complete_results}.

\subsection{Results for CHAIN DoubleDQN in MinAtar}
\label{subsec:exp_vb}

We use DoubleDQN (DDQN)~\citep{HasseltGS16DoubleDQN} as the value-based method and MinAtar~\citep{Young2019MinAtar} as the experiment environments.
MinAtar is an Atari-inspired testbed for convenient evaluation and reproduction.
We build our DoubleDQN based on the official MinAtar code with no change to the network structure and hyperparameters.
We implement CHAIN DDQN by adding a few lines of code to apply the value churn reduction regularization in the standard training of the $Q$-network (Eq.~\ref{eq:q_objectives}).
For CHAIN DDQN, $\lambda_{Q}$ is set to $50$ for Breakout and $100$ for the other tasks.
For CHAIN DDQN with automatic adjustment of $\lambda_Q$ (denoted by the suffix \textbf{`Auto'}), the target relative loss scale $\beta$ is set to $0.05$ for all the tasks.

\begin{wrapfigure}{r}{0cm}
\begin{minipage}[t]{.45\linewidth}
\centering
\vspace{-0.5cm}
\hspace{-0.1cm}
\scalebox{0.5}{
\includegraphics[width=0.95\textwidth]
{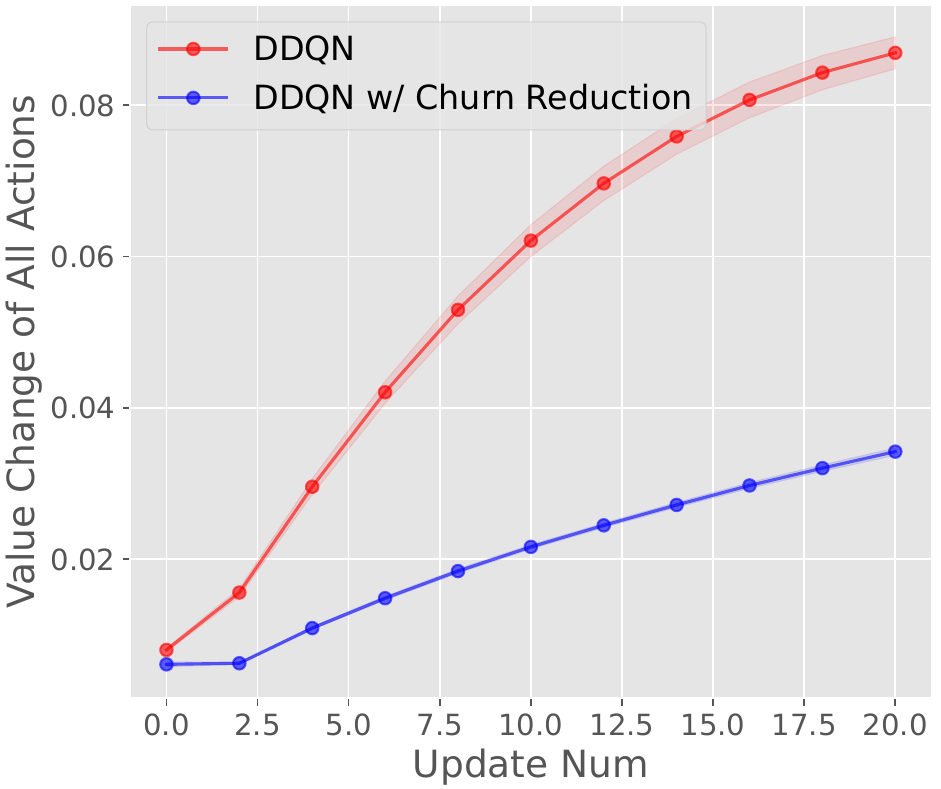}
\includegraphics[width=0.95\textwidth]
{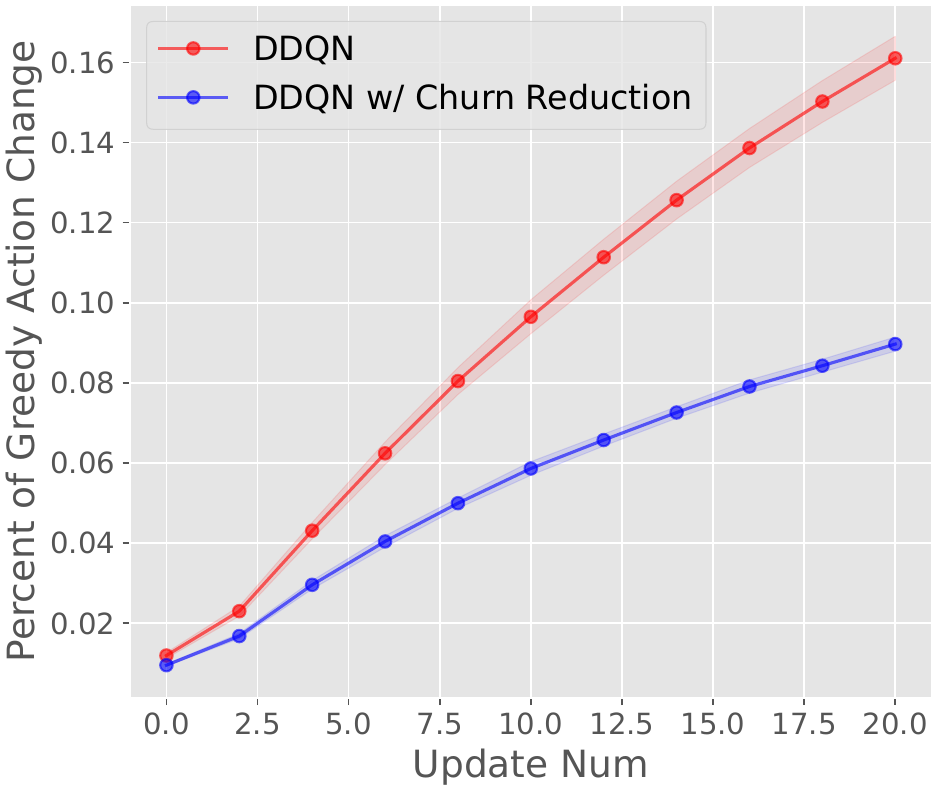}
}
\vspace{-0.2cm}
\caption{The value churn (\textit{left}) and the greedy action deviation percentage (\textit{right})
in Breakout w/ and w/o CHAIN.
}
\label{figure:stats_breakout}
\vspace{-0.4cm}
\end{minipage}
\end{wrapfigure}

First, to answer Question (1), we report the value churn and the greedy action deviation of DDQN in Figure~\ref{figure:stats_breakout}.
Each point means the average metric across randomly sampled states and the whole learning process.
As expected, we can observe that the value churn accumulates as more training updates take place, leading to the growth of greedy action deviation.
With CHAIN, the churn and deviation are reduced significantly.
We refer the readers to Figure~\ref{figure:minatar_agg_stats} for more statistics on the value churn.

\begin{figure*}[h]
\vspace{-0.4cm}
\begin{center}
\includegraphics[width=1.0\textwidth]{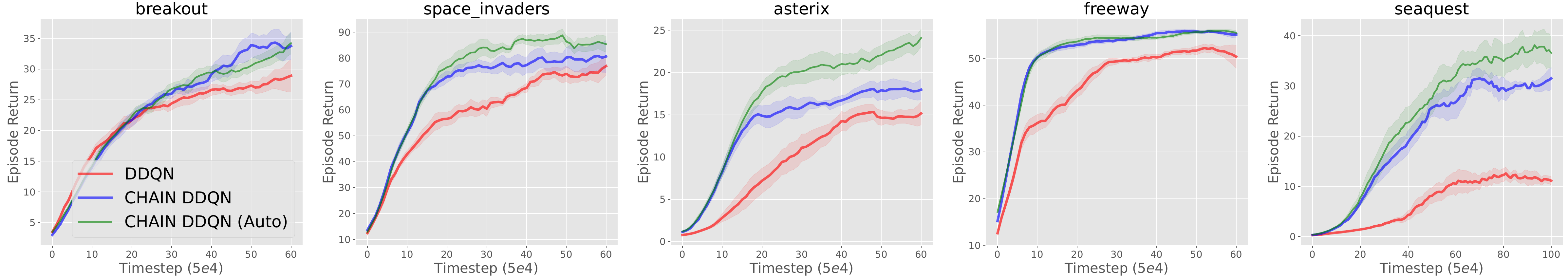}
\end{center}
\vspace{-0.4cm}
\caption{The evaluation of \methodName DoubleDQN in MinAtar regarding episode return.
Curves and shades denote means and standard errors over six random seeds.
}
\vspace{-0.3cm}
\label{figure:minatar_auto_eval}
\end{figure*}

Further, we show the learning curves of CHAIN DDQN regarding episode return in Figure~\ref{figure:minatar_auto_eval}.
We can see that \methodName
consistently achieves clear improvements over DDQN in terms of both sample efficiency and final scores, especially
for Asterix and Seaquest.
Moreover, CHAIN (Auto) matches or surpasses the results achieved by manual coefficients,
which supports Question (2) positively.
In the next subsection, we evaluate how much CHAIN can improve policy gradient-based RL algorithms.

\subsection{Results for CHAIN PPO in MuJoCo and DMC}
\label{subsec:exp_pg}

Corresponding to the second DRL scenario discussed in~\Cref{subsec:consequences_of_churn}, we focus on the policy churn in Proximal Policy Optimization (PPO)~\citep{SchulmanWDRK17PPO} and try to answer the first three questions for policy gradient-based RL algorithms.
We build the experiments on the public implementation of PPO from \texttt{CleanRL}~\citep{huang2022cleanrl}
and use the continuous control tasks in MuJoCo and DeepMind Control (DMC) as the environments for evaluation.
Following the same principle, we implement CHAIN PPO by adding the policy churn reduction regularization to the standard PPO policy training (Eq.~\ref{eq:l_pc}), with no other modification to the public PPO implementation.

First, to understand the level of churn, we compare PPO and CHAIN PPO with different choices of $\lambda_{\pi}$ in terms of policy churn and episode return.
In summary, we observed that PPO also exhibits clear policy churn, and CHAIN significantly reduces it throughout learning, which answers Question (1).
Figure~\ref{figure:chain_ppo_results_full} shows the details for this analysis.
Note that more policy churn makes it more likely to violate the trust region as $\mathcal{C}_{\pi} \propto r(\phi,\phi^{\prime})$ discussed in \Cref{subsec:consequences_of_churn}.
In turn, we also observed CHAIN PPO consistently outperforms PPO in Ant-v4 and HalfCheetah-v4 across different choices of $\lambda_{\pi}$.

\begin{figure*}[h]
\vspace{-0.3cm}
\begin{center}
\includegraphics[width=1.0\textwidth]{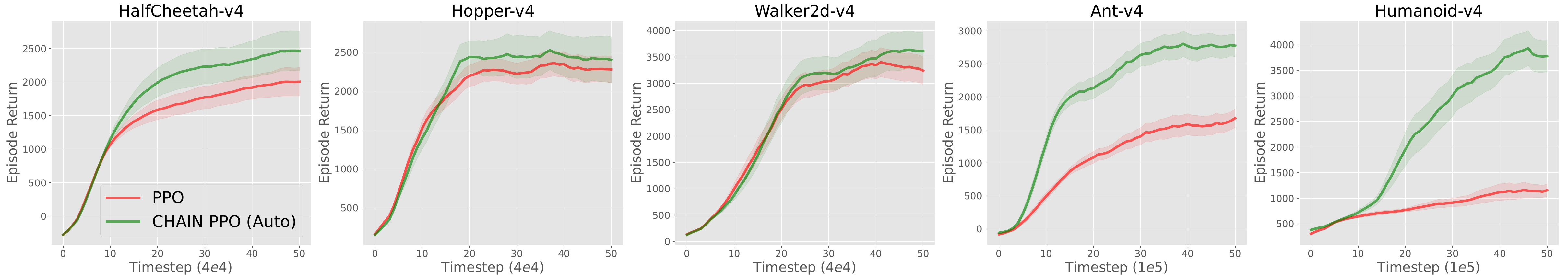}
\includegraphics[width=1.0\textwidth]{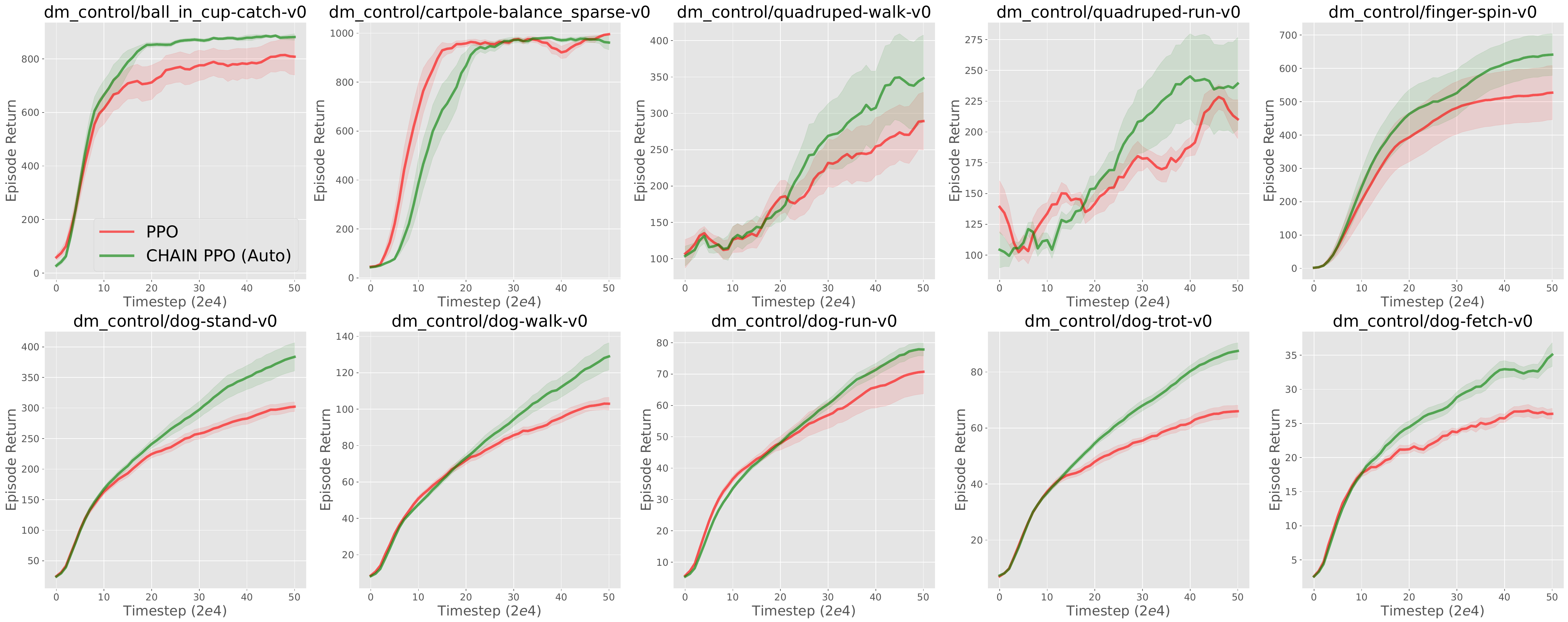}
\end{center}
\vspace{-0.4cm}
\caption{
The evaluation of \methodName PPO in MuJoCo and DeepMind Control (DMC) tasks regarding episode return.
Curves and shades denote means and standard errors over twelve random seeds.
}
\vspace{-0.3cm}
\label{figure:mujoco_and_dmc_auto_eval}
\end{figure*}

Further, we aim to answer Question (2) and evaluate whether CHAIN can improve the learning performance of PPO in terms of episode return.
For CHAIN PPO (Auto), we set the target relative loss scale $\beta$ to 0.1 for MuJoCo tasks and 0.02 for DMC tasks.
The results for MuJoCo and DMC tasks are reported in Figure~\ref{figure:mujoco_and_dmc_auto_eval}.
The results show that CHAIN PPO outperforms PPO in most cases with higher sample efficiency and final episode return, often significantly. 
We believe that our results reveal a promising direction to improve more trust-region-based and constraint-based methods in DRL by addressing the issues caused by churn.

\subsection{Results for Deep Actor-Critic Methods with CHAIN in MuJoCo and D4RL}
\label{subsec:exp_deep_ac}

Next, we continue our empirical study and evaluate our method for deep actor-critic (AC) methods. 
We separate our study into online and offline settings, as presented below.

\noindent \textbf{Online AC methods}
We use TD3~\citep{Fujimoto2018TD3} and SAC~\citep{HaarnojaZAL18SAC} and MuJoCo environments based on the public implementation of TD3 and SAC from \texttt{CleanRL}.
Since AC methods are bilaterally affected by churn, we consider two variants of CHAIN, either of which only applies the value churn reduction (VCR) or the policy churn reduction (PCR).

\begin{figure*}[h]
\vspace{-0.3cm}
\begin{center}
\includegraphics[width=0.975\textwidth]{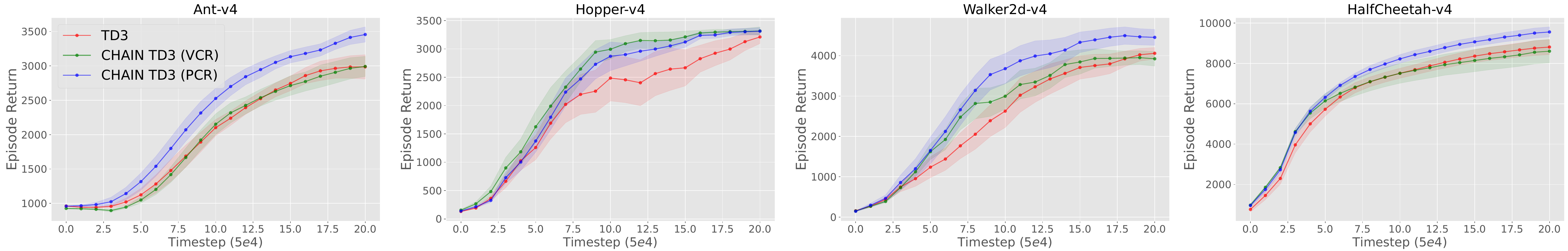}
\includegraphics[width=0.975\textwidth]{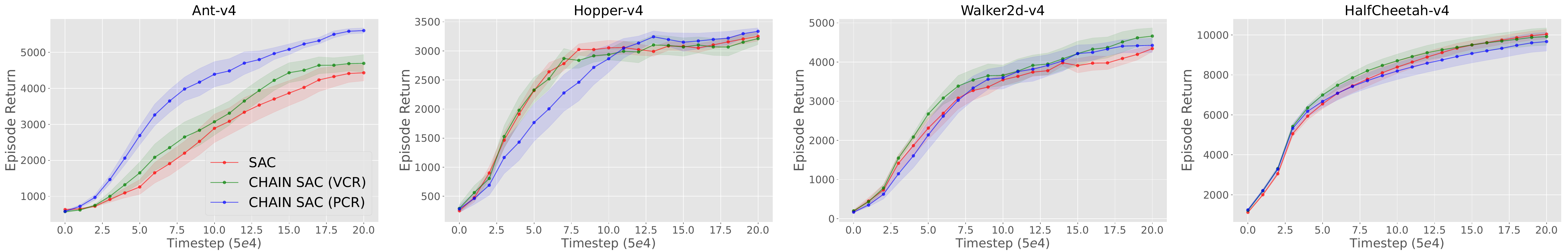}
\end{center}
\vspace{-0.4cm}
\caption{The evaluation of \methodName TD3 and \methodName SAC in MuJoCo regarding episode return.
}
\vspace{-0.3cm}
\label{figure:mujoco_eval}
\end{figure*}

For Question (1), we again find that both TD3 and SAC exhibit value and policy churn in all environments, and CHAIN-VCR and CHAIN-PCR effectively reduce them respectively in Figure~\ref{figure:mujoco_td3_churn_reduction} and~\ref{figure:mujoco_sac_churn_reduction}.
For Question (2), Figure~\ref{figure:mujoco_eval} shows the evaluation results regarding episode return.
We can see that CHAIN-PCR often improves the learning performance, especially for Ant-v4; in contrast, CHAIN-VCR improves slightly.
We hypothesize that this is because the policy interacts with the environment directly, and the target critic-network also helps to reduce the value churn due to its delayed synchronization with the online critic.

Due to the limitation of space, 
we refer the readers to Appendix~\ref{app:more_results_chain_ac} for
more results on churn reduction, the influence of different choices of $\lambda_{Q},\lambda_{\pi}$, the results of combining VCR and PCR, as well as the effect of auto-adjustment of the regularization coefficient for TD3 and SAC.

\noindent \textbf{Offline AC methods}
In Offline RL, a policy is trained over a fixed dataset. We investigate if reducing churn can also improve the convergence of Offline RL.
We use IQL~\citep{KostrikovNL22IQL} with D4RL Antmaze dataset~\citep{Fu2020D4RL} and AWAC~\citep{Nair2020AWAC} with Adroit for our demonstration due to their popularity and good performance in corresponding tasks. 
Concretely, we use the public implementation and benchmark scores for IQL and AWAC from \texttt{CORL}\footnote{\url{https://github.com/tinkoff-ai/CORL}}.
To apply CHAIN to IQL and AWAC,
we implement the regularization for value churn reduction (VCR, Eq.~\ref{eq:q_objectives}) and policy churn reduction (PCR, Eq.~\ref{eq:p_objectives}) separately by adding a couple of lines of code without any other modification. 
We use $\lambda_{\pi} = 1e3$ for both CHAIN IQL (PCR) and CHAIN AWAC (PCR); $\lambda_{Q} = 0.01$ for CHAIN IQL (VCR) and $0.1$ for CHAIN AWAC (VCR) across different tasks. 
The results are summarized in Table~\ref{table:chain_offline_iql} and~\ref{table:chain_offline_awac}.

\begin{table}[h]
  \vspace{-0.1cm}
  \caption{Results for CHAIN IQL in Antmaze, with means and standard errors over twelve seeds.}
  \vspace{-0.1cm}
  \centering
  \scalebox{0.8}{
  \begin{tabular}{c|c|c|c}
    \toprule
    Task & IQL & CHAIN IQL (PCR) & CHAIN IQL (VCR) \\
    \midrule
    AM-umaze-v2 & 77.00 $\pm$ 5.52 & \textbf{84.44 $\pm$ 3.19} & 83.33 $\pm$ 2.72 \\
    AM-umaze-diverse-v2 & 54.25 $\pm$ 5.54 & 62.50 $\pm$ 3.75 & \textbf{71.67 $\pm$ 7.23} \\
    AM-medium-play-v2 & 65.75 $\pm$ 11.71 & \textbf{72.50 $\pm$ 2.92} & 70.00 $\pm$ 3.33 \\
    AM-medium-diverse-v2 & 73.75 $\pm$ 5.45 & \textbf{76.67 $\pm$ 4.51} & 66.67 $\pm$ 3.79 \\
    AM-large-play-v2 & 42.00 $\pm$ 4.53 & \textbf{50.00 $\pm$ 4.56} & 43.33 $\pm$ 4.14\\
    AM-large-diverse-v2 & 30.25 $\pm$ 3.63 & 26.67 $\pm$ 3.96 & \textbf{31.67 $\pm$ 2.31} \\
    \bottomrule
  \end{tabular}
  }
\label{table:chain_offline_iql}
\vspace{-0.4cm}
\end{table}
\begin{table}[h]
\caption{Results for CHAIN AWAC in Adroit, with means and standard errors over twelve seeds.
}
\vspace{-0.1cm}
\centering
\scalebox{0.8}{
\begin{tabular}{c|c|c|c}
    \toprule
    Task & AWAC & CHAIN AWAC (PCR) & CHAIN AWAC (VCR) \\
    \midrule
    pen-human-v1 & 81.12 $\pm$ 13.47 & \textbf{99.72 $\pm$ 2.04} & 97.37 $\pm$ 3.51 \\
    pen-cloned-v1 & 89.56 $\pm$ 15.57 & 95.49 $\pm$ 2.34 & \textbf{96.66 $\pm$ 2.54} \\
    \bottomrule
  \end{tabular}
}
\label{table:chain_offline_awac}
\vspace{-0.2cm}
\end{table}

We observe that both CHAIN PCR and CHAIN VCR  improve the scores for IQL and AWAC in most Antmaze and Adroit tasks.
We hypothesize that CHAIN suppresses churn in the training of value and policy networks, thus reducing the bias caused by churn in parameter updates.
One thing here that differs from TD3 and SAC considered in the online setting is that the policy network of IQL has no impact on the training of the value networks since the value networks (i.e., $Q$ and $V$) are trained purely based on in-sample data without accessing $a^{\prime} = \pi_{\phi}(s^{\prime})$. Thus, although IQL does not exhibit a chain effect explicitly,
the policy and value networks of IQL still have churns, which are reduced by \methodName in this case.
We provide a further empirical study in Appendix~\ref{app:more_for_iql}.

\subsection{Scaling DRL Agents with CHAIN}
\label{subsec:exp_scaling}

It is widely known that scaling DRL agents up is challenging. Naively scaling DRL agents by widening or deepening the conventional MLP networks straightforwardly often fails and could even lead to collapse.
Here, we investigate the relationship between churn and scale for DRL agents, as well as the effect of CHAIN on boosting scaling performance, to answer Question (3).
We take PPO and MuJoCo tasks as the exemplary setting and scale up both the policy and value networks by a scale-up ratio within $\{2, 4, 8, 16\}$ via \textit{widening} or \textit{deepening}. Note that the default network architecture (i.e., when the scale-up ratio equals one) for both the policy and value networks is a two-layer MLP with $256$ neurons for each layer, followed by an output layer.

\begin{figure*}[h]
\vspace{-0.3cm}
\begin{center}
\includegraphics[width=0.85\textwidth]{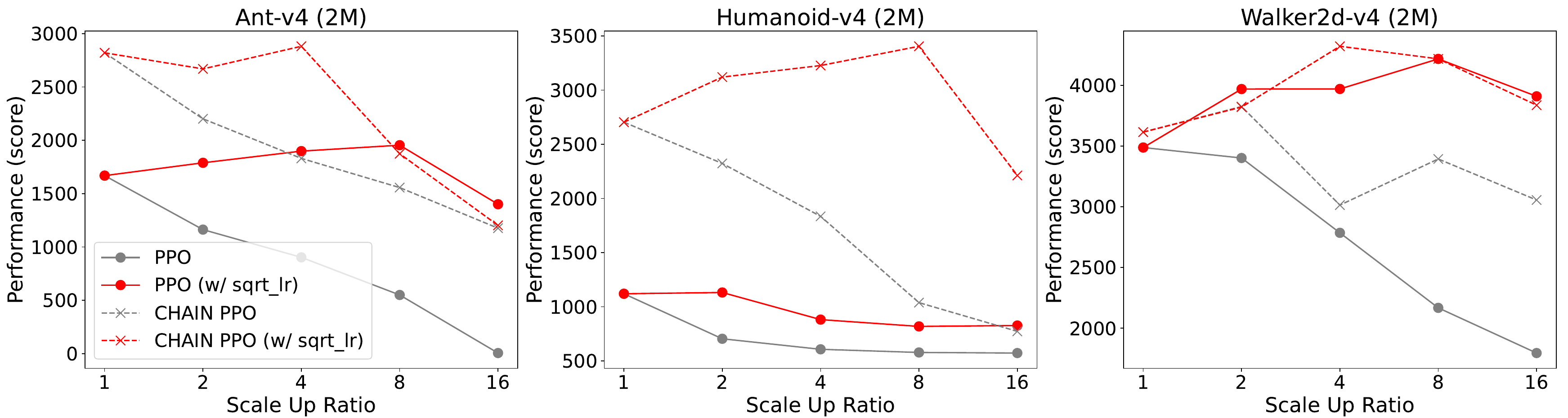}
\end{center}
\vspace{-0.4cm}
\caption{The results regarding episode return for scaling PPO via widening. CHAIN helps to scale almost across all the configurations. Similar results can be found for widening scaling in Figure~\ref{figure:ppo_scaling_pcr_full}.}
\vspace{-0.2cm}
\label{figure:ppo_scaling_pcr}
\end{figure*}

As expected, we observed that the performance of PPO degraded severely as the scale-up ratio increased,
as shown by the solid gray lines in Figure~\ref{figure:ppo_scaling_pcr}.
Inspired by the prior study~\citep{Ob24Pruned}, we found using a decreased learning rate as \texttt{lr / sqrt(scale-up ratio)} alleviates the degradation of PPO scaling to some degree (shown by the solid red lines).
We then use the learning rate setting below by default.
From the perspective of churn, we also observed that scaling PPO escalates the scale of the policy churn in PPO.
More results can be found in Appendix~\ref{app:more_for_ppo_scaling}.
Therefore, we then evaluate the effect of CHAIN in this scaling setting.
The results are shown in Figure~\ref{figure:ppo_scaling_pcr} with dashed lines.
By comparing the lines in the same color, we found that CHAIN improves the learning performance of PPO across almost all scale-up ratios and the two learning rate settings.

In addition, we extend the training horizon from 2M to 10M for Ant, Humanoid, and Walker2d.
The results are reported in Table~\ref{table:ppo_scaling_with_chain}.
For both widening or deepening cases, CHAIN helps to scale up PPO and achieves clear improvement in terms of episode return. Comparatively, scaling by widening slightly outperforms deepening, which echoes the observation in ~\citep{OtaOJMN20Canincreasing} to some extent. 

\begin{table}[h]
\vspace{-0.2cm}
\caption{Scaling PPO with CHAIN. Means and standard errors of final episode return over six seeds. 
  }
  \vspace{-0.1cm}
  \centering
  \scalebox{0.75}{
  \begin{tabular}{c|ccc}
    \toprule
    Alg. (scale)   & Ant (10M) & Human. (10M) & Walker2d (10M) \\
    \midrule
    PPO  & 2238.45  $\pm$  256.07 & 1620.45  $\pm$  212.10 & 3316.77  $\pm$  269.42 \\
    \midrule
    PPO (4x \underline{wider}) & 3013.95 $\pm$  223.77 & 2998.18  $\pm$  237.95 & 3795.05 $\pm$  208.17  \\
    CHAIN PPO (4x \underline{wider})  & \textbf{4916.66}  $\pm$  109.01 & \textbf{4830.58} $\pm$  231.42 & \textbf{4668.16} $\pm$  234.45 \\
    \midrule
    PPO (4x \underline{deeper}) & 2777.62  $\pm$  136.78 & 3489.47  $\pm$  166.28 & 2845.68  $\pm$  260.02 \\
    CHAIN PPO (4x \underline{deeper})  & \textbf{3760.46}  $\pm$  158.29 & \textbf{4090.80}  $\pm$  187.01 & \textbf{3242.83}  $\pm$  173.54 \\
    \midrule
    PPO (8x \underline{wider}) & 4235.63  $\pm$  209.68 & 3198.01  $\pm$  344.71 & 4335.17  $\pm$  123.12 \\
    CHAIN PPO (8x \underline{wider})  & \textbf{5592.94}  $\pm$  205.23 & \textbf{5246.15}  $\pm$  130.73 & \textbf{5161.42} $\pm$  343.40 \\
    \midrule
    PPO (8x \underline{deeper}) & 3364.46 $\pm$  173.25 & 2780.67  $\pm$  304.10 & 3057.89  $\pm$  299.61 \\
    CHAIN PPO (8x \underline{deeper})  & \textbf{4278.76}  $\pm$  122.61 & \textbf{4434.48}  $\pm$  144.92 & \textbf{3324.29}  $\pm$  255.88 \\
    \bottomrule
  \end{tabular}
  }
\label{table:ppo_scaling_with_chain}
\vspace{-0.2cm}
\end{table}

Our results indicate that uncontrolled churn could be a possible reason for the scaling issue of DRL agents, and CHAIN improves scaling by reducing churn effectively.
Though appealing, CHAIN does not fully address the scaling issue per se, and achieves sub-linear scaling on DRL agents.

\section{Conclusion}
\label{sec:conclusion}

In this paper, we conduct a formal study of churn in a general view and present
the chain effect of value and policy churn.
The chain effect indicates a compounding cycle, which biases parameter updates throughout learning.
We propose an easy-to-implement method for value and policy churn reduction. Our experimental results demonstrate the effectiveness of our method in reducing churn and improving learning performance over a range of DRL environments and algorithms.

\section*{Acknowledgements}
We want to acknowledge funding support from NSERC, FQRNT and CIFAR and compute support from Digital Research Alliance of Canada, Mila IDT, and NVidia.

\bibliography{references}
\bibliographystyle{plainnat}

\newpage
\appendix

\section{Limitations}
\label{app:limitations}
Our work is limited in several directions below and we expect further studies on these points in the future.
\begin{itemize}
    \item First, the theoretical analysis of the chain effect under concrete assumptions remains to be explored.
    Other perspectives that may influence churn, such as network structure, representation learning, and experience replay are not considered in this work, which are worthwhile to study in the future.
    \item Moreover, although we provided a simple method to adjust the regularization coefficients dynamically throughout learning by keeping a consistent relative loss scale, it is not sufficient for us to use the same hyperparameter for different domains. We believe that using normalization techniques to unify the scales in different domains is necessary to address this point, similar to the work in \citep{Dreamerv3}.
    \item Besides, for algorithms that involve the learning of both policy and value (e.g., deep AC methods), the implementation of \methodName in different problems faces the question of choosing the best option among PCR, VCR and DCR. For this, we expect to develop a better method to integrate the effect of PCR and VCR in the future.
    \item Another remaining problem is the lack of an in-depth understanding of the positive and negative effects of churn on the generalization of DRL agents, which could drive new methods that better leverage the potential of churn.
\end{itemize}

\section{Additional Formal Analysis}
\label{app:derivations}

\subsection{The NTK Expressions of Two Types of Deviation}

The NTK expression for the action gradient deviation $\mathcal{D}^{Q}_{\nabla_{a}}$ is straightforward to obtain by plugging in the NTK expression for the $Q$-value churn:
\begin{equation*}
\begin{aligned}
    \mathcal{D}^{Q}_{\nabla_{a}}(\theta,\theta^{\prime}) & = \nabla_{\bar a} Q_{\theta^{\prime}}(\bar s, \bar a)|_{\bar a=\pi(s)} - \nabla_{\bar a} Q_{\theta}(\bar s,\bar a)|_{\bar a=\pi(\bar s)} \\
    & = \nabla_{\bar a} (\underbrace{Q_{\theta^{\prime}}(\bar s, \bar a) - Q_{\theta}(\bar s, \bar a)}_{\mathcal{C}_{Q}(\theta, \theta^{\prime})})|_{\bar a=\pi(\bar s)} \\
    & \approx \nabla_{\bar a} (k_{\theta}(\bar s, \bar a,s,a) \delta_{\theta}(s,a))|_{\bar a=\pi(\bar s)} \\
    & \approx \left(\frac{\partial^2 Q_{\theta}(\bar s, \bar a)}{\partial \theta \partial \bar a}\right) \Delta_{\theta}|_{\bar a=\pi(\bar s)} 
\end{aligned}
\end{equation*}

And for the policy value deviation $\mathcal{D}^{\pi}_{Q}$, the NTK expression is obtained by performing Taylor expansion of $\pi_{\phi}(\bar s)$ and plugging in the NTK expression for the policy churn:
\begin{equation*}
\begin{aligned}
    \mathcal{D}^{\pi}_{Q} & (\phi, \phi^{\prime}) 
    = (\nabla_{\bar a} Q_{\theta}(\bar s, \bar a)|_{\bar a=\pi_{\phi}(\bar s)})^{\top}(\underbrace{\pi_{\phi^{\prime}}(\bar s) - \pi_{\phi}(\bar s)}_{\mathcal{C}_{\pi}(\phi, \phi^{\prime})}) + O(\| \underbrace{\pi_{\phi^{\prime}}(\bar s) - \pi_{\phi}(\bar s)}_{\mathcal{C}_{\pi}(\phi, \phi^{\prime})} \|^2) \\
    & \approx (\nabla_{\bar a} Q_{\theta} (\bar s, \bar a) |_{\bar a=\pi_{\phi}(\bar s)})^{\top} k_{\phi}(\bar s, s) \nabla_{a}Q_{\theta}(s,a)|_{a=\pi_{\phi}(s)} \\
    & \approx (\underbrace{\nabla_{\phi}\pi_{\phi} (\bar s) \nabla_{\bar a} Q_{\theta}(\bar s, \bar a)|_{\bar a=\pi_{\phi}(\bar s)}}_{\text{DPG of } \pi_{\phi} \ \text{at } \bar s})^{\top} \Delta_{\phi} \\
\end{aligned}
\end{equation*}

The NTK expression of the two types of deviation above indicates that action gradient deviation and policy value deviation are mainly influenced by the second-order partial derivatives of $Q_{\theta}$ and the deterministic policy gradient (DPG) of $\pi_{\phi}$ at the state $\bar s$ considered, in addition to the parameter update.
An interesting observation here is, policy value deviation of policy churn has an \textit{implicit optimization} effect (while altered by $\Delta_{\phi}$) for the reference states (i.e., $\bar s$) although the parameter update $\phi \rightarrow \phi^{\prime}$ is performed for the training states (i.e., $s$).
We observe that $\mathcal{D}^{\pi}_{Q}$ is positive in overall in our empirical investigation later (Appendix~\ref{app:complete_results}).
However, careful considerations are needed in the future because the implicit optimization could be delusional as churn also occurs in $Q_{\theta}$ through the learning process.

\subsection{Complete Discussions and Derivations for Section~\ref{subsec:chain_effect}}
\label{app:chain_effect_detail}

Recall conventional gradients $\Delta_{\theta}, \Delta_{\phi}$ of value network and policy network:
\begin{equation*}
\begin{aligned}
    & \Delta_{\theta}(s,a)  =  \nabla_{\theta}Q_{\theta}(s,a)  \delta_{\theta}(s,a) \\
    & \Delta_{\phi}(s)  = \nabla_{\phi}\pi_{\phi}(s) \nabla_{a}  Q_{\theta}(s,a) |_{a=\pi_{\phi}(s)}
\end{aligned}
\end{equation*}
where $\delta_{\theta}(s,a) = r + \gamma Q_{\theta}(s^{\prime}, \pi_{\phi}(s^{\prime}))$.

Now let us re-consider the gradients by taking into consideration the value and policy churn, denoted by $\tilde{\Delta}_{\theta}, \tilde{\Delta}_{\phi}$.
Our purpose is to formulate the difference between $\tilde{\Delta}_{\theta}, \tilde{\Delta}_{\phi}$ and $\Delta_{\theta}, \Delta_{\phi}$ as functions of the policy and value churns, as well as their derivatives. 

\paragraph{More derivatives of $\mathcal{C}_Q, \mathcal{C}_{\pi}$} Before looking into the gradients $\tilde{\Delta}_{\theta}, \tilde{\Delta}_{\phi}$, we need three more definitions for network parameter gradient deviation caused by value and policy churn, and action gradient deviation caused by policy churn, during parameter update $\theta \rightarrow \theta^{\prime}, \phi \rightarrow \phi^{\prime}$.
\begin{itemize}
    \item $Q$-network Gradient Deviation of Value Churn: $\mathcal{D}^{Q}_{\nabla_{\theta}}(\theta, \theta^{\prime}, \{\bar s, \bar a \}) = \nabla_{\theta^{\prime}}Q_{\theta^{\prime}}(\bar s, \bar a) - \nabla_{\theta}Q_{\theta}(\bar s, \bar a)$.
    \item Policy Network Gradient Deviation of Policy Churn: $\mathcal{D}^{\pi}_{\nabla_{\phi}}(\phi, \phi^{\prime}, \{\bar s \}) = \nabla_{\phi^{\prime}}\pi_{\phi^{\prime}}(\bar s) - \nabla_{\phi}\pi_{\phi}(\bar s)$.
    \item Action Gradient Deviation of Policy Churn: $\mathcal{D}^{\pi}_{\nabla_{a}}(\phi, \phi^{\prime}, \{ \bar s \}) = \nabla_{\bar a^{\prime}} Q(\bar s, \bar a^{\prime})|_{\bar a^{\prime}=\pi_{\phi^{\prime}}(\bar s)}- \nabla_{\bar a} Q(\bar s, \bar a)_{\bar a = \pi_{\phi}(\bar s)}$
\end{itemize} 
Note $\mathcal{D}^{\pi}_{\nabla_{a}}(\phi, \phi^{\prime}, \{ s \})$ denotes the action gradient caused by the policy churn for the $Q$ function affected by the churns $\uwave{Q_{\theta}}$ rather than $Q_{\theta}$. $\mathcal{D}^{\pi}_{\nabla_{a}}$ is a further consequence of policy value deviation caused by the policy churn $\mathcal{D}^{\pi}_{Q}(\phi^{-}, \phi, \{ s \})$.

\begin{figure}
\begin{center}
\includegraphics[width=0.55\textwidth]{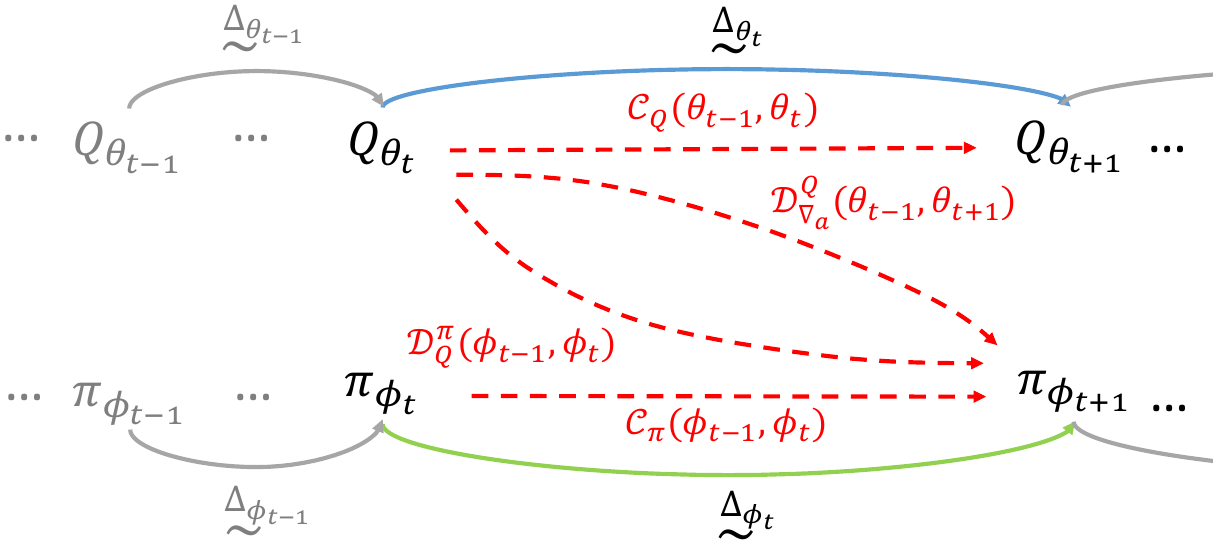}
\end{center}
\vspace{-0.3cm}
\caption{An illustration of the parameter update chain with the value and policy churn. 
Blue and green arrows denote iterative updates of the value network and policy network. The value and policy churn and their deviations are marked in red.
}
\vspace{-0.2cm}
\label{figure:update_chain}
\end{figure}

To shed light on the long-term effect of churn,
we depict a typical iterative update scenario in Figure~\ref{figure:update_chain}, where the $Q$-network and policy network update with corresponding gradients in a chain.
Different from conventional analysis, we explicitly consider the value and policy churn and study how they affect the chain of parameter updates.

As in~\cref{subsec:chain_effect}, we focus on the segment $(\theta^{-}, \phi^{-}) \rightarrow (\theta, \phi) \rightarrow (\theta^{\prime}, \phi^{\prime})$ on the chain of update.
The churns occurred during the past update $(\theta^{-}, \phi^{-}) \rightarrow (\theta, \phi)$, which should further affect the update about to perform at present $(\theta, \phi) \rightarrow (\theta^{\prime}, \phi^{\prime})$.
Concretely, the churns and the deviations affect the following aspects:
\textbf{(1)} $Q$-value estimate and \textbf{(2)} action selection in both TD error and policy objective,
and \textbf{(3)} the gradient of network parameter.
Now we are ready to deduce the parameter update under the effect of the value and policy churn below.
Note that we use $\uwave{ \ \ }$ under the terms that are affected by value and policy churn:
\begin{equation*}
\begin{aligned}
    & \tilde{\Delta}_{\theta}(s,a)  =  \uwave{\nabla_{\theta }Q_{\theta}}(s,a)  \tilde{\Delta}_{\theta}(s,a)  =  \uwave{\nabla_{\theta}Q_{\theta}}(s,a)  ( r + \gamma \uwave{Q_{\theta}}(s^{\prime}, \uwave{a}^{\prime}) - \uwave{Q_{\theta}}(s,a)) \\
    & =   \Big( \nabla_{\theta}Q_{\theta}(s,a) + ( \uwave{\nabla_{\theta }Q_{\theta}}(s,a) - \nabla_{\theta}Q_{\theta}(s,a)) \Big) \Big( r + \gamma \Big[ Q_{\theta}(s^{\prime}, a^{\prime}) + ( \uwave{Q_{\theta}}(s^{\prime}, \uwave{a}^{\prime}) - \uwave{Q_{\theta}}(s^{\prime}, a^{\prime})) \\ 
    & + (\uwave{Q_{\theta}}(s^{\prime}, a^{\prime}) - Q_{\theta}(s^{\prime}, a^{\prime})) \Big ] - \Big[ Q_{\theta}(s, a) + (\uwave{Q_{\theta}}(s,a) - Q_{\theta}(s, a)) \Big] \Big) \\
    & =   \left( \nabla_{\theta} Q_{\theta}(s,a) + \textcolor{red}{\mathcal{D}^{Q}_{\nabla_{\theta}}(\theta^{-}, \theta, \{s,a \} )} \right) \Big( \delta_{\theta}(s,a) + \gamma ( \textcolor{red}{\mathcal{D}^{\pi}_{Q} (\phi^{-}, \phi, \{s^{\prime}, \uwave{\pi_{\phi}}(s^{\prime})\}) + \mathcal{C}_{Q}(\theta^{-}, \theta, \{s^{\prime}, \pi_{\phi}(s^{\prime}) \} ) }) \\
    & - \textcolor{red}{\mathcal{C}_{Q}(\theta^{-}, \theta, \{s, a\})} \Big) \\
    & = \nabla_{\theta} Q_{\theta}(s,a) \delta_{\theta}(s,a) \\
    & + \gamma \nabla_{\theta} Q_{\theta}(s,a) (\textcolor{red}{\mathcal{D}^{\pi}_{Q} (\phi^{-}, \phi, \{s^{\prime}, \uwave{\pi_{\phi}}(s^{\prime})\}) + \mathcal{C}_{Q}(\theta^{-}, \theta, \{s^{\prime}, \pi_{\phi}(s^{\prime}) \} ) } ) - \nabla_{\theta} Q_{\theta}(s,a) \textcolor{red}{\mathcal{C}_{Q}(\theta^{-}, \theta, \{s, a\})} \\
    & + \textcolor{red}{\mathcal{D}^{Q}_{\nabla_{\theta}}(\theta^{-}, \theta, \{s,a \} )} \Big( \delta_{\theta}(s,a) + \gamma ( \textcolor{red}{\mathcal{D}^{\pi}_{Q} (\phi^{-}, \phi, \{s^{\prime}, \uwave{\pi_{\phi}}(s^{\prime})\}) + \mathcal{C}_{Q}(\theta^{-}, \theta, \{s^{\prime}, \pi_{\phi}(s^{\prime}) \} ) }) - \textcolor{red}{\mathcal{C}_{Q}(\theta^{-}, \theta, \{s, a\})} \Big) \\
\end{aligned}
\end{equation*}
\begin{equation*}
\begin{aligned}
    & \tilde{\Delta}_{\phi}(s)  = \uwave{\nabla_{\phi}\pi_{\phi}}(s) \uwave{\nabla_{a} Q_{\theta}}(s, \uwave{a}) |_{\uwave{a=\pi_{\phi}}(s)} \\
    & = \Big( \nabla_{\phi}\pi_{\phi}(s) + (\uwave{\nabla_{\phi}\pi_{\phi}}(s) - \nabla_{\phi}\pi_{\phi}(s)) \Big) 
    \Big( \nabla_{a} Q_{\theta}(s,a) + (\uwave{\nabla_{a} Q_{\theta}}(s, \uwave{a}) - \nabla_{a} \uwave{Q_{\theta}}(s,a)) \\
    & + (\nabla_{a} \uwave{Q_{\theta}}(s,a) - \nabla_{a} Q_{\theta}(s,a))  \Big)|_{\uwave{a=\pi_{\phi}}(s), a=\pi_{\phi}(s)} \\
    & = \left( \nabla_{\phi}\pi_{\phi}(s) + \textcolor{red}{\mathcal{D}^{\pi}_{\nabla_{\phi}}(\phi^{-}, \phi)} \right) 
    \left( \nabla_{a} Q_{\theta}(s,a) + \textcolor{red}{\mathcal{D}^{\pi}_{\nabla_{a}}(\phi^{-}, \phi, \{ s \}) + \mathcal{D}^{Q}_{\nabla_{a}}(\theta^{-}, \theta, \{ s, a \})}  \right)|_{ a=\pi_{\phi}(s)} \\
    & = \nabla_{\phi}\pi_{\phi}(s)\nabla_{a} Q_{\theta}(s,a)|_{a=\pi_{\phi}(s)} + \nabla_{\phi}\pi_{\phi}(s) \left(\textcolor{red}{\mathcal{D}^{\pi}_{\nabla_{a}}(\phi^{-}, \phi, \{ s \}) + \mathcal{D}^{Q}_{\nabla_{a}}(\theta^{-}, \theta, \{ s, a \})}  \right)|_{a=\pi_{\phi}(s)} \\
    & + \textcolor{red}{\mathcal{D}^{\pi}_{\nabla_{\phi}}(\phi^{-}, \phi)}
    \left( \nabla_{a} Q_{\theta}(s,a) + \textcolor{red}{\mathcal{D}^{\pi}_{\nabla_{a}}(\phi^{-}, \phi, \{ s \}) + \mathcal{D}^{Q}_{\nabla_{a}}(\theta^{-}, \theta, \{ s, a \})}  \right)|_{a=\pi_{\phi}(s)} \\
\end{aligned}
\end{equation*}
As a result, we can find that the value and policy churn, as well as the deviations derived, introduce biases in the parameter updates.

Recall that the parameter updates cause the churns constantly, 
the analysis on the update segment $(\theta^{-}, \phi^{-}) \rightarrow (\theta, \phi) \rightarrow (\theta^{\prime}, \phi^{\prime})$
can be forwarded and leads to the cycle illustrated in Figure~\ref{figure:vicious_cycle}:
(1) parameter update causes the value and policy churn $\mathcal{C}_{Q},\mathcal{C}_{\pi}$, which (2) further deviates the action gradient and policy value $\mathcal{D}^{\pi}_{Q},\mathcal{D}^{Q}_{\nabla_{a}}$ (and the other deviations);
(3) the churns and the deviations then bias following parameter updates with $\tilde{\Delta}_{\theta}, \tilde{\Delta}_{\phi}$.
Consequently, the value and policy churn and the parameter update bias \textit{accumulate} and can \textit{amplify each other} throughout the learning process.

Apparently, the long-term chain effect is intricate as 
the training process is stochastic and the churns are influenced by various factors, e.g., network structure, learning objective, and data distribution.
We leave theoretical studies under concrete assumptions in the future.

\section{Experimental Details}
\label{app:exp_details}

\subsection{Compute Resources and Time Cost}
We use Nvidia V100 GPU for our experiments.
The additional computation introduced by CHAIN lies in the sampling and training with a second batch of data for the churn reduction regularization.
In essence, it just increases the batch size by 2 to use CHAIN, resulting in a constant change in complexity with respect to the default DRL algorithm. This is a small price to pay to reduce the churn and increase policy performance. In practice, we also observe similar wall-clock time cost for standard DRL methods and their CHAIN versions.

More importantly, since CHAIN often brings higher sample efficiency, i.e., achieving the same level of score with fewer interaction steps, this implies that CHAIN accelerates learning and achieves good performance earlier. Concrete examples can be found in our experimental results.

\subsection{Empirical Metrics for the Investigation of the Chain Effect}
\label{app:metrics_empirical_investigation}

To investigate the extent of the value churn and the policy churn, we compare the output changes between current networks $(\theta_{t},\phi_{t})$ and past network versions $(\theta^{-},\phi^{-}) \in \{(\theta_{t-i},\phi_{t-i})\}_{i=1}^{N}$.

For policy-based methods, we study TD3 and SAC for MuJoCo.
We compute the value churn (\textcolor{blue}{$\hat{\mathcal{C}}_{Q_{sa}},\hat{\mathcal{C}}_{|Q_{sa}|}$}), the policy churn (\textcolor{blue}{$\hat{\mathcal{C}}_{\pi}$}) and the value deviation of policy churn (\textcolor{blue}{$\hat{\mathcal{D}}^{\pi}_{Q}$}) throughout learning.

For value-based methods, we compute the percentage of the greedy action deviation (\textcolor{blue}{$\hat{\mathcal{D}}^{Q}_{a^{\star}}$}), the value churn of greedy action (\textcolor{blue}{$\hat{\mathcal{C}}_{Q_{a^{\star}}}$}) and the value churn of all actions (\textcolor{blue}{$\hat{\mathcal{C}}_{Q_{s}},\hat{\mathcal{C}}_{|Q_{s}|}$})
These metrics are summarized in Table~\ref{table:churn_metrics}.
\begin{table}[ht]
  \centering
  \caption{Churn and deviation metrics used in our experiments. $\theta_{t},\phi_{t}$ and $\theta^{-}, \phi^{-}$ are current networks and previous networks. The metrics are averaged over $\bar s, \bar a$ in a reference buffer.
  }
  \scalebox{0.9}{
  \begin{tabular}{c|c}
    \toprule
    \multicolumn{2}{c}{The Metrics for Policy-based Methods}\\
    \midrule
    \textcolor{blue}{$\hat{\mathcal{C}}_{Q_{sa}}(\theta^{-}, \theta_{t})$} & $ Q_{\theta_{t}}(\bar s, \bar a) - Q_{\theta^{-}}(\bar s, \bar a) $ \ \text{(for TD3, SAC)} \\
    \textcolor{blue}{$\hat{\mathcal{C}}_{|Q_{sa}|}(\theta^{-}, \theta_{t})$} & $ | Q_{\theta_{t}}(\bar s, \bar a) - Q_{\theta^{-}}(\bar s, \bar a) | $ \ \text{(for TD3, SAC)} \\
    \multirow{2}{*}{\textcolor{blue}{$\hat{\mathcal{C}}_{\pi}(\phi^{-}, \phi_{t})$}} & $ \| \pi_{\phi_{t}}(\bar s) - \pi_{\phi^{-}}(\bar s)) \|_1 $ \ \text{(for TD3, PPO\footnote{We do not use a distribution metric because the conventional PPO implementation uses state-independent variance parameters.})} \\
    & $ \text{KL}(\pi_{\phi_{t}}(\cdot|\bar s), \pi_{\phi^{-}}(\cdot|\bar s))$ \ \text{(for SAC)} \\
    \textcolor{blue}{$\hat{\mathcal{D}}^{\pi}_{Q}(\phi^{-}, \phi_{t})$} & $Q_{\theta_{t}}(\bar s, \pi_{\phi_{t}}(\bar s)) - Q_{\theta_{t}}(\bar s, \pi_{\phi^{-}}(\bar s))$\\
    \midrule
    \multicolumn{2}{c}{The Metrics for Value-based Methods}\\
    \midrule
    \textcolor{blue}{$\hat{\mathcal{C}}_{|Q_{s}|}(\theta^{-}, \theta_{t})$} & $\frac{1}{|A|} \sum_{a \in A} | Q_{\theta_{t}}(\bar s, a) - Q_{\theta^{-}}(\bar s, a) | $ \\
    \textcolor{blue}{$\hat{\mathcal{C}}_{Q_{a^{\star}}}(\theta^{-}, \theta_{t})$} & $\max_{a^{\prime}} Q_{\theta_{t}}(\bar s, a^{\prime}) - \max_a Q_{\theta^{-}}(\bar s, a)$\\
    \textcolor{blue}{$\hat{\mathcal{D}}^{Q}_{a^{\star}}(\theta^{-}, \theta_{t})$} & $\mathbb{I}_{\{\arg\max_a Q_{\theta^{-}}(\bar s, a)\}} (\arg\max_{a^{\prime}} Q_{\theta_{t}}(\bar s, a^{\prime}))$\\
    \bottomrule
  \end{tabular}
  }
  \vspace{-0.15cm}
\label{table:churn_metrics}
\end{table}

We use $N=50$ and $N=20$ for MuJoCo and MinAtar environments and
compute the metrics at an interval of 1k parameter updates. 
For each type of metric in~\Cref{table:churn_metrics}
and update number $i \in \{1,\dots, N\}$,
we compute the mean of the quantities throughout learning.
We refer the readers to Figure~\ref{figure:minatar_agg_stats},
\ref{figure:mujoco_td3_churn_reduction} and
\ref{figure:mujoco_sac_churn_reduction} for the results in Appendix~\ref{app:complete_results}.

\subsection{Code Implementation}

We use the public implementations of PPO, TD3 and SAC in \texttt{CleanRL}\footnote{\url{https://github.com/vwxyzjn/cleanrl}} as our codebase.
The actor and critic networks are two-layer MLPs with 256 units for each layer.
For DoubleDQN, we modified the DQN implementation provided in the official code of MinAtar paper\footnote{\url{https://github.com/kenjyoung/MinAtar}} with no change to the network structure, recommended hyperparameter choices, etc.
Hyperparameters are listed in Table~\ref{table:minatar_hyperparm_common}, Table~\ref{table:ppo_mujoco_hyperparameter} and Table~\ref{table:mujoco_hyperparameter}.
For the experiments in the offline RL setting,
we use the public implementation and benchmark scores for IQL and AWAC from \texttt{CORL}\footnote{\url{https://github.com/tinkoff-ai/CORL}}.

One thing to note is that the data in the training batch and the regularization batch should be non-overlapping in principle, but we found simply sampling two random batches from the replay buffer independently works well.
From the probabilistic perspective, the overlap could happen at a low probability, which is determined by the batch size and the size of the replay buffer.

We make no change to the state/observation and reward of MuJoCo and MinAtar environments.
No additional tricks like state normalization, reward normalization are used in our experiments.

\subsection{Other Discussions}

\paragraph{Discussion on the Data Batches Used in Training and Anlysis}

We use separate sets of data for churn reduction regularization (i.e., the regularization set) and churn investigation/evaluation (i.e., the actual reference set), and they are randomly sampled at each network update. In other words, if count in the regular batch for training, we have three separate batches in total: (1) a regular training batch for standard RL loss computation, (2) a regularization batch for churn reduction loss, and (3) a reference batch for churn evaluation (optional) throughout the learning process. 
Note that the reference batch is only used for churn evaluation and does not influence learning.

\begin{table}[ht]
  \caption{Hyperparameters of DoubleDQN used in MinAtar environments. The values of conventional hyperparameters are taken from the recommended values  in~\citep{Young2019MinAtar}.
  }
  \centering
  \scalebox{0.95}{
  \begin{tabular}{c|c}
    \toprule
    \multicolumn{2}{c}{DoubleDQN Hyperparameters}\\
    \midrule
    Learning Rate & 3e$^{-4}$\\
    Training Interval & 1 step \\
    Discount Factor ($\gamma$) & 0.99 \\
    Hard Replacement Interval & 1000 steps \\
    Replay Buffer Size & 0.5M \\
    Batch Size & 32 \\
    Initial $\epsilon$ & 1.0 \\
    End $\epsilon$ & 0.1 \\
    $\epsilon$ Decay Steps & 0.5M \\
    Initial Random Steps & 10k \\
    \midrule
    \multicolumn{2}{c}{Value Churn Reduction Hyperparameter}\\
    \midrule
    \multirow{2}{*}{Value Regularization Coefficient ($\lambda_{Q}$)}
     & 50 for Breakout \\
    & 100 for others \\
    \midrule
    Target Relative Loss $\beta$ for Auto $\lambda_Q$ & 0.05 \\
    \bottomrule
  \end{tabular}
  }
\label{table:minatar_hyperparm_common}
\end{table}

\begin{table}[ht]
  \caption{Hyperparameters of PPO used in MuJoCo environments. The values of conventional hyperparameters are taken from the recommended values in \texttt{CleanRL}.
  }
  \centering
  \scalebox{0.95}{
  \begin{tabular}{c|c}
    \toprule
    \multicolumn{2}{c}{PPO Hyperparameters}\\
    \midrule
    Learning Rate & 3e$^{-4}$\\
    Training Interval & 2048 steps \\
    Discount Factor ($\gamma$) & 0.99 \\
    GAE Parameter ($\lambda$) & 0.95 \\
    Num. of Minibatches & 32 \\
    Update Epoch & 10 \\
    \multirow{2}{*}{Clipping Range Parameter ($\epsilon$)}
     & 0.1 for MuJoCo tasks \\
    & 0.2 for DMC tasks \\
    \midrule
    \multicolumn{2}{c}{Policy Churn Reduction Hyperparameter}\\
    \midrule
    \multirow{2}{*}{Policy Regularization Coefficient ($\lambda_{\pi}$)}
     & 5000 for Ant-v4 \\
    & 50 for other MuJoCo tasks \\
    \midrule
    \multirow{2}{*}{Target Relative Loss $\beta$ for Auto $\lambda_{\pi}$}
     & 0.1 for MuJoCo tasks \\
    & 0.02 for DMC tasks \\
    \bottomrule
  \end{tabular}
  }
\label{table:ppo_mujoco_hyperparameter}
\end{table}

\begin{table}[ht]
  \caption{Hyperparameters of TD3 and SAC used in MuJoCo environments. The values of conventional hyperparameters are taken from the recommended values in \texttt{CleanRL}. '-' means 'not applicable'.
  }
  \centering
  \scalebox{0.95}{
  \begin{tabular}{c|c|c}
    \toprule
    \multicolumn{3}{c}{TD3 \& SAC Hyperparameters}\\
    \midrule
    Hyperparameters & TD3 & SAC \\
    \midrule
    Actor Learning Rate & 3e$^{-4}$ & 3e$^{-4}$\\
    Critic Learning Rate & 3e$^{-4}$ & 3e$^{-4}$\\
    Actor Training Interval & 2 steps & 1 step \\
    Critic Training Interval & 1 step & 1 step\\
    Exploration Noise & $\mathcal{N}(0,0.1)$ & - \\
    Target Action Noise & $\mathcal{N}(0,0.2)$ & - \\
    Target Action Noise Clip & 0.5 & - \\
    Discount Factor ($\gamma$) & 0.99 & 0.99 \\
    Soft Replacement Ratio & 0.005 & 0.005 \\
    Initial Random Steps & 5k & 5k \\
    Replay Buffer Size & 1M & 1M\\
    Batch Size & 256 & 256 \\
    Optimizer & Adam & Adam  \\
    \midrule
    \multicolumn{3}{c}{Churn Reduction Hyperparameters (refer to the study in Fig.~\ref{figure:mujoco_eval}, \ref{figure:mujoco_td3_churn_reduction}, \ref{figure:mujoco_sac_churn_reduction})}\\
    \midrule
    Policy Regularization Coefficient ($\lambda_{\pi}$) & $\{0.1, 1, 20\}$ & $\{1e^{-4}, 5e^{-4}, 1e^{-3}\}$ \\
    Value Regularization Coefficient ($\lambda_{Q}$) & \multicolumn{2}{c}{$\{0.1, 0.5, 1.0\}$}\\
    \midrule
    Target Relative Loss $\beta$ for Auto $\lambda_{\pi}$
     & \multicolumn{2}{c}{5e$^{-5}$} \\
    \bottomrule
  \end{tabular}
  }
\label{table:mujoco_hyperparameter}
\end{table}

\begin{algorithm}
    \begin{algorithmic}[1]
        \STATE Initialize the policy network $\pi_{\phi}$ and the value network $Q_{\theta}$ with parameters $\phi, \theta$ if they exist
        \STATE Initialize an empty buffer $D$
        \STATE \textcolor{blue}{set churn reduction hyperparamters $\lambda_{Q}, \lambda_{\pi}$}
        \STATE \textcolor{blue}{(Optional) Set target relative loss scale $\beta$ for auto-adjustment of $\lambda_{Q}, \lambda_{\pi}$}
        \FOR{iteration $t  =  1, 2, 3, ...$}
            \STATE \textcolor{gray}{// Interact with the environment and collect samples}
            \STATE Rollout the policy with $\pi_{\phi}$ or $Q_{\theta}$ and store interaction data in $D$
            \STATE \textcolor{gray}{// Perform value and policy network learning}
            \IF{time to update}
                \STATE Sample a training batch $B_{\text{train}}$ and
                \textcolor{blue}{a reference batch $B_{\text{ref}}$} from $D$
                \STATE Update $Q_{\theta}$ with $L(\theta, B_{\text{train}})$, \textcolor{blue}{$L_{\text{QC}}(\theta, B_{\text{ref}})$} by Eq.~\ref{eq:l_qc},~\ref{eq:q_objectives}
    	    \STATE Update $\pi_{\phi}$ with $L(\phi, B_{\text{train}})$, \textcolor{blue}{$L_{\text{PC}}(\phi, B_{\text{ref}})$} by Eq.~\ref{eq:l_pc},~\ref{eq:p_objectives}
                \STATE \textcolor{blue}{(Optional) Re-calculate $\lambda_{Q}, \lambda_{\pi}$ according to $\beta$ (refer to Section~\ref{subsec:chain_method})}
            \ENDIF
        \ENDFOR
        \end{algorithmic}
    \caption{
    Deep RL with Churn Approximated ReductIoN (\textbf{\methodName}).
    }
\label{alg:drl_with_chain}
\end{algorithm}

\clearpage
\section{Complete Results}
\label{app:complete_results}

\subsection{More Empirical Analysis on the Value Churn in DoubleDQN}

\begin{figure*}[ht]
\begin{center}
\subfigure[Percentage of Greedy Action Change]{
\includegraphics[width=1.0\textwidth]{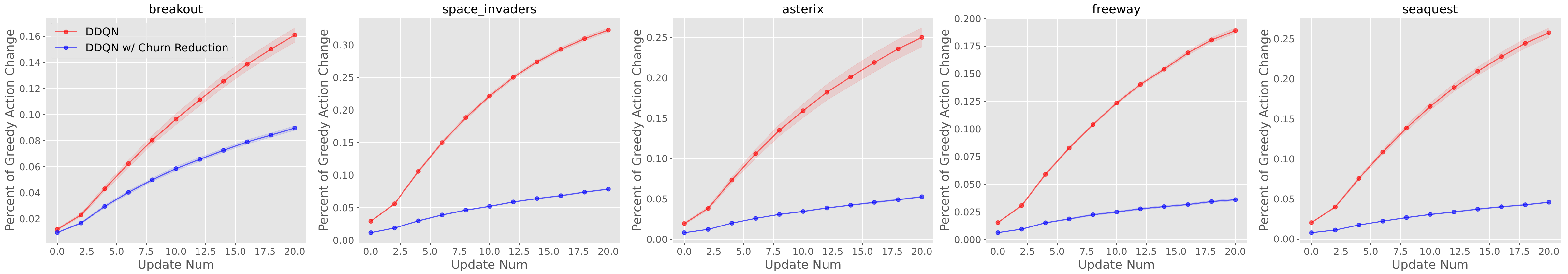}
\label{subfig:minatar_agg_greedy_a_diff}
}\\
\vspace{-0.2cm}
\subfigure[Value Change of Greedy Action]{
\includegraphics[width=1.0\textwidth]{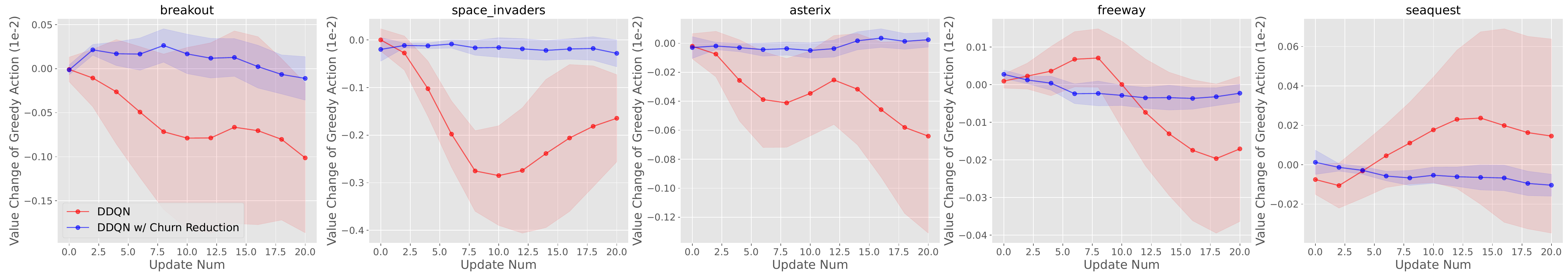}
\label{subfig:minatar_agg_maxq_diff}
}\\
\vspace{-0.2cm}
\subfigure[Value Change of All Actions]{
\includegraphics[width=1.0\textwidth]{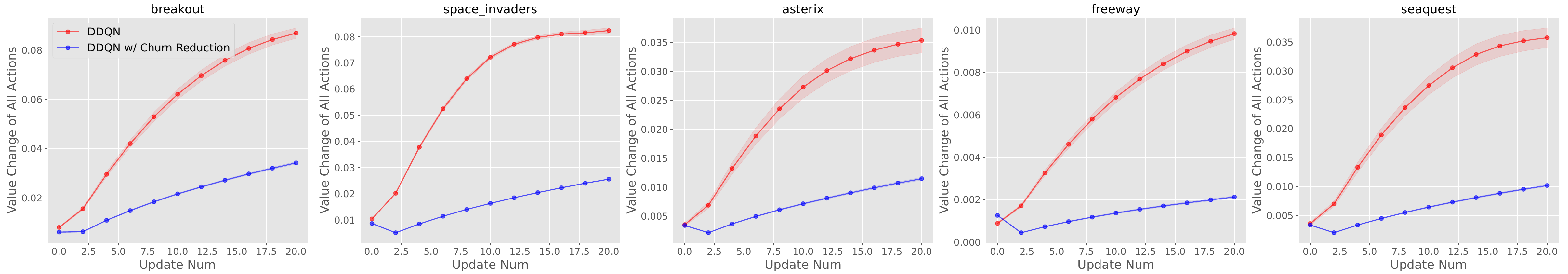}
\label{subfig:minatar_agg_q_diff_abs}
}
\end{center}
\vspace{-0.3cm}
\caption{Different statistics on value churn of DoubleDQN in MinAtar. Horizontal axes are numbers of parameter updates after which the statistics are computed. Each point is obtained by averaging all the quantities throughout learning. Curves and shades denote means and standard errors across six random seeds.
}
\label{figure:minatar_agg_stats}
\end{figure*}

In Figure~\ref{figure:minatar_agg_stats} shows the statistics of the three metrics defined in Table~\ref{table:churn_metrics}.
Horizontal axes show the number of parameter updates after which the statistics are computed. Each point is obtained by averaging all the quantities throughout learning. Curves and shades denote means and standard errors across six random seeds.

We can observe that the amount of value churn (\textcolor{blue}{$\hat{\mathcal{C}}_{|Q_{s}|}$}) accumulates as the update number increases. Although there does not exist an explicit policy network, the percentage of greedy action deviation (\textcolor{blue}{$\hat{\mathcal{C}}_{Q_{a^{\star}}}$}) goes up to over $20\%$.
However, different from the case for TD3 and SAC, the value of greedy action (\textcolor{blue}{$\hat{\mathcal{D}}^{Q}_{a^{\star}}$}) decreases and the value churn (\textcolor{blue}{$\hat{\mathcal{C}}_{Q_{s}}$}) exhibits a similar trend.

\subsection{More Results of CHAIN PPO}
\label{app:more_results_chain_ppo}

\begin{figure*}[ht]
\begin{center}
\subfigure[Ant]{
\includegraphics[width=1.0\textwidth]{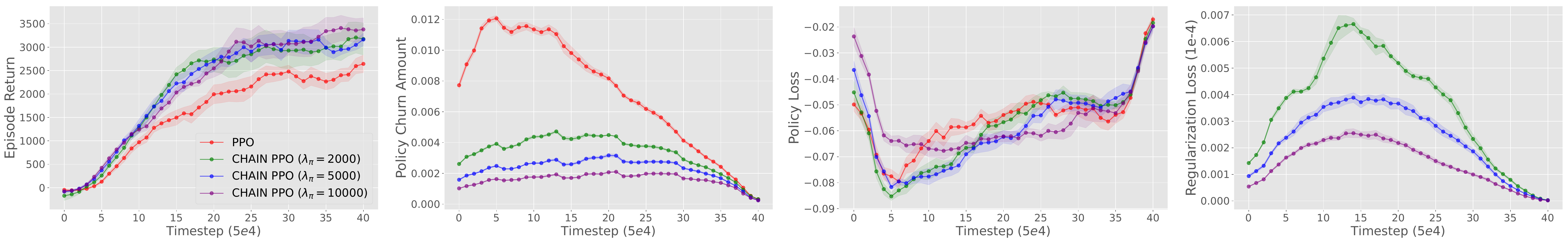}
}\\
\vspace{-0.2cm}
\subfigure[HalfCheetah]{
\includegraphics[width=1.0\textwidth]{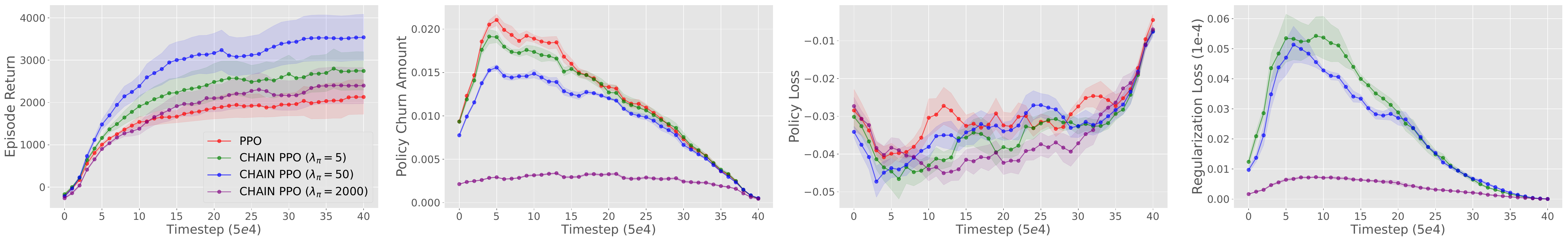}
}
\vspace{-0.2cm}
\subfigure[Walker2d]{
\includegraphics[width=1.0\textwidth]{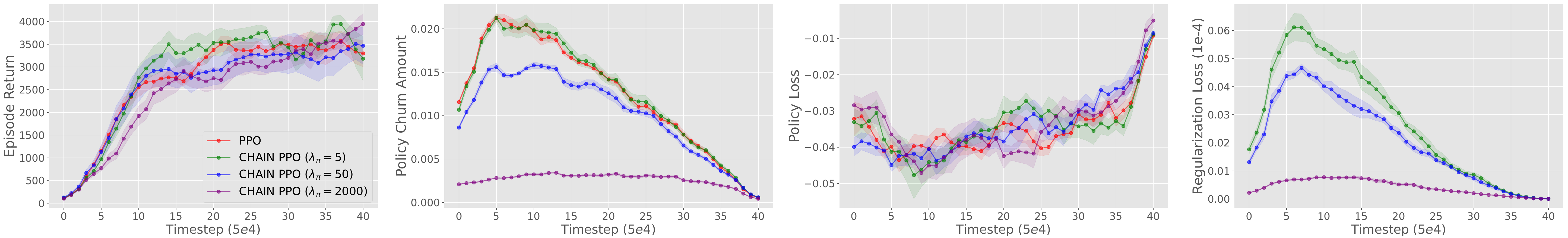}
}
\subfigure[Hopper]{
\includegraphics[width=1.0\textwidth]{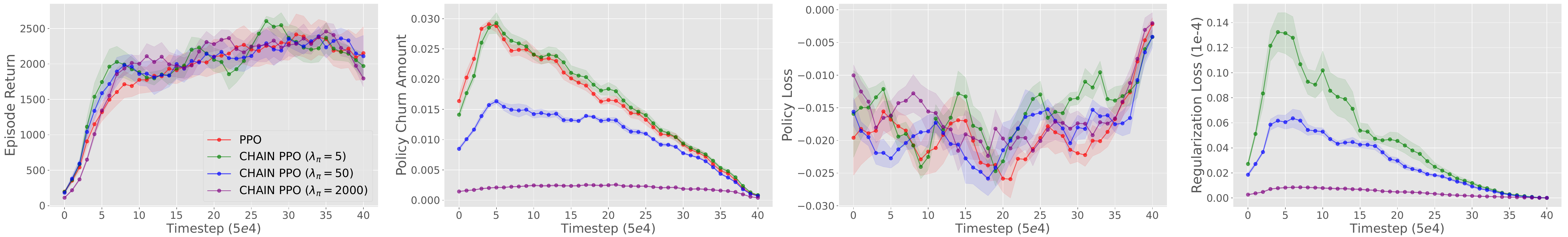}
}
\end{center}
\vspace{-0.2cm}
\caption{Results for CHAIN PPO, including the learning performance, the amount of the policy churn, the loss of conventional policy training, and the loss of regularization.
}
\label{figure:chain_ppo_results_full}
\end{figure*}

Figure~\ref{figure:chain_ppo_results_full} provides the learning performance of CHAIN PPO on four MuJoCo tasks.
Moreover, we also report the conventional PPO policy loss and the regularization loss $L_{\text{PC}}(\phi)$ during the learning process.
Figure~\ref{figure:chain_ppo_results_full} also shows the results for different choices of the hyperparameter $\lambda_{\pi}$.

From the results, we can observe:
(1) PPO exhibits clear policy churn (the second column).
One thing to note is that the fading of the policy churn is due to the decay of the learning rate in \texttt{CleanRL}'s PPO implementation.
CHAIN (2) reduces the policy churn of PPO effectively, and (3) shows performance improvement in Ant and HalfCheetah and comparable performance in Hopper and Walker2d.
For the hyperparameter choice, in practice, we found the churn reduction coefficient between 1e3 and 1e4 works well for Ant, while 50 works best in HalfCheetah.

We hypothesize
that the choice of $\lambda_{\pi}$ is likely to be related to the scale difference between policy loss $-J(\phi)$ (the third column) and regularization term $L_{\text{PC}}(\phi)$ (the fourth column).
This motivates the proposal of our method for automatic adjustment of the regularization coefficient presented in Section~\ref{subsec:chain_method}.

\subsection{More Results of CHAIN for Deep AC Methods}
\label{app:more_results_chain_ac}

\begin{figure*}[ht]
\begin{center}
\subfigure[HalfCheetah-v4]{
\includegraphics[width=1.0\textwidth]{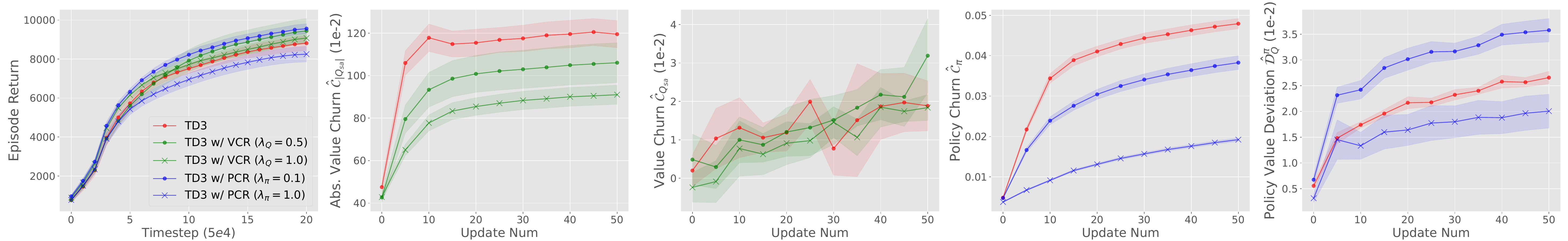}
}\\
\vspace{-0.2cm}
\subfigure[Hopper-v4]{
\includegraphics[width=1.0\textwidth]{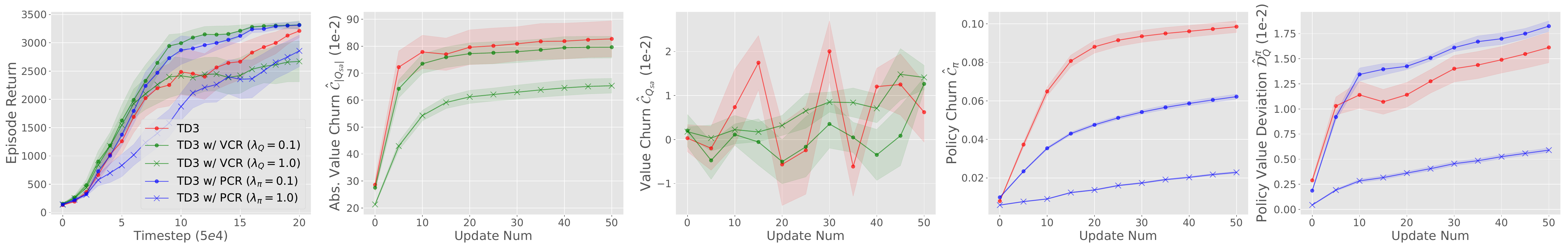}
}\\
\vspace{-0.2cm}
\subfigure[Walker2d-v4]{
\includegraphics[width=1.0\textwidth]{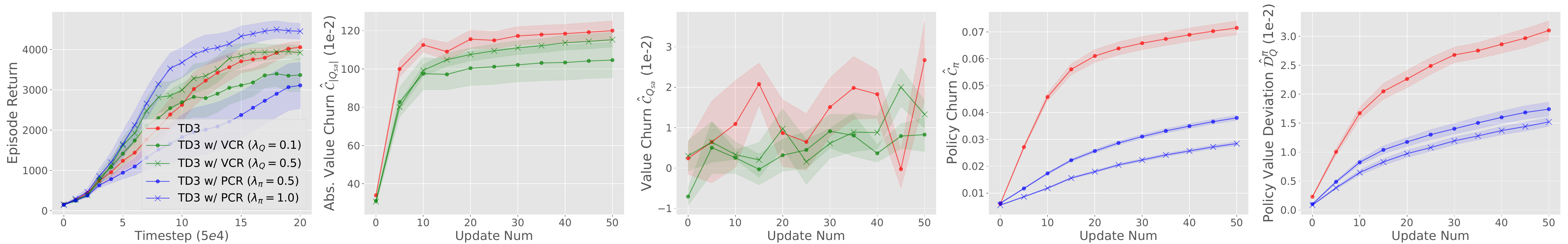}
}\\
\vspace{-0.2cm}
\subfigure[Ant-v4]{
\includegraphics[width=1.0\textwidth]{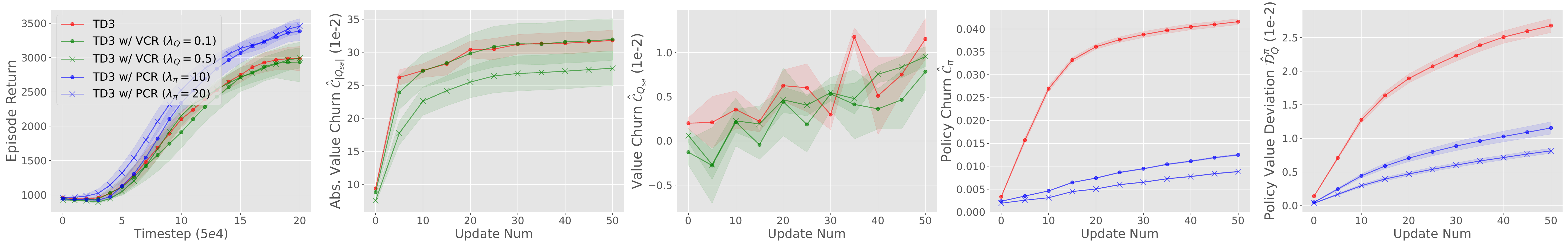}
}
\end{center}
\vspace{-0.3cm}
\caption{Churn reduction of TD3 in MuJoCo. 
The 2\textit{nd}-4\textit{th} columns report the value of the four metrics defined in Table~\ref{table:churn_metrics}.
Horizontal axes are numbers of parameter updates after which the statistics are computed. Each point is obtained by averaging all the quantities throughout learning.
}
\label{figure:mujoco_td3_churn_reduction}
\end{figure*}

\begin{figure*}[ht]
\begin{center}
\subfigure[HalfCheetah-v4]{
\includegraphics[width=1.0\textwidth]{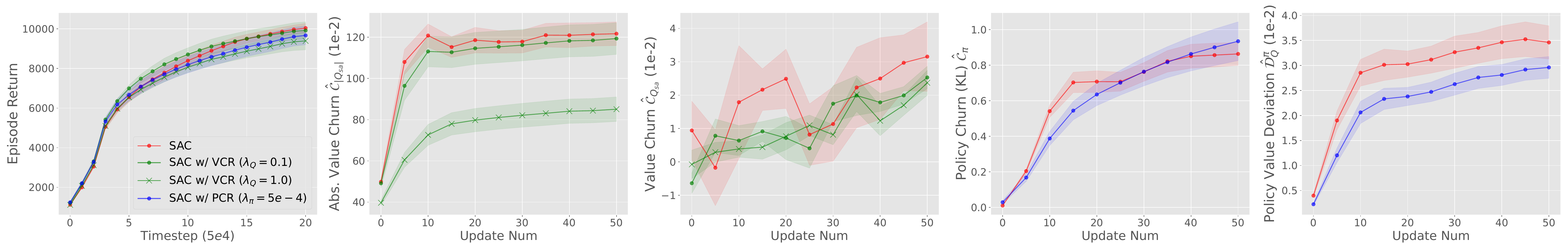}
}\\
\vspace{-0.2cm}
\subfigure[Hopper-v4]{
\includegraphics[width=1.0\textwidth]{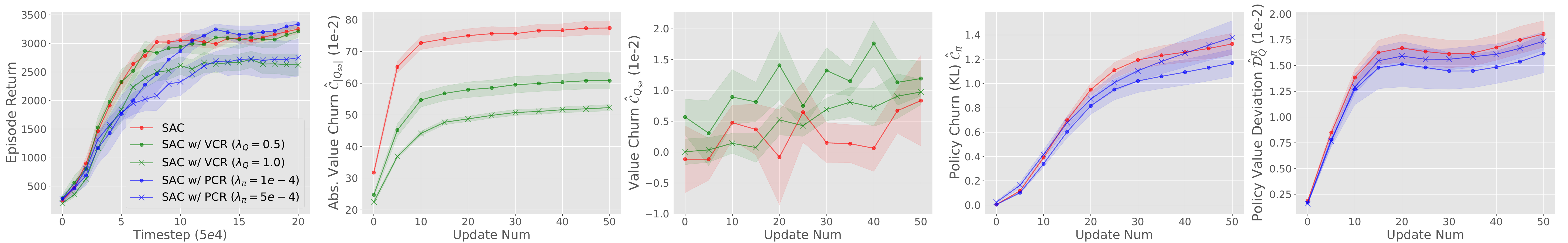}
}\\
\vspace{-0.2cm}
\subfigure[Walker2d-v4]{
\includegraphics[width=1.0\textwidth]{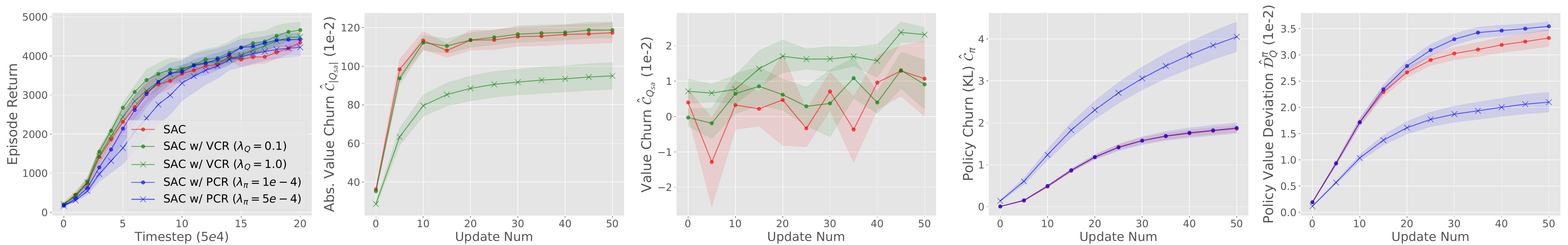}
}\\
\vspace{-0.2cm}
\subfigure[Ant-v4]{
\includegraphics[width=1.0\textwidth]{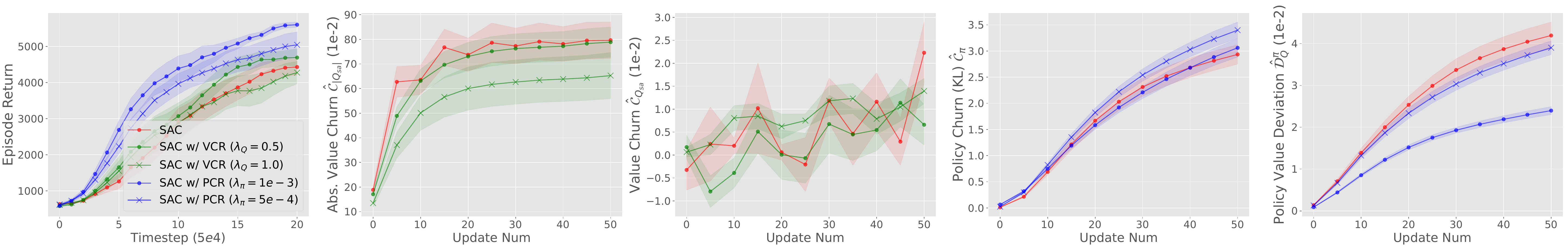}
}
\end{center}
\vspace{-0.3cm}
\caption{Churn reduction of SAC in MuJoCo. 
The 2\textit{nd}-4\textit{th} columns report the value of the four metrics defined in Table~\ref{table:churn_metrics}.
Horizontal axes are numbers of parameter update after which the statistics are computed. Each point is obtained by averaging all the quantities throughout learning.
}
\label{figure:mujoco_sac_churn_reduction}
\end{figure*}

\begin{figure*}[ht]
\begin{center}
\subfigure[Value Churn Reduction (VCR)]{
\includegraphics[width=1.0\textwidth]{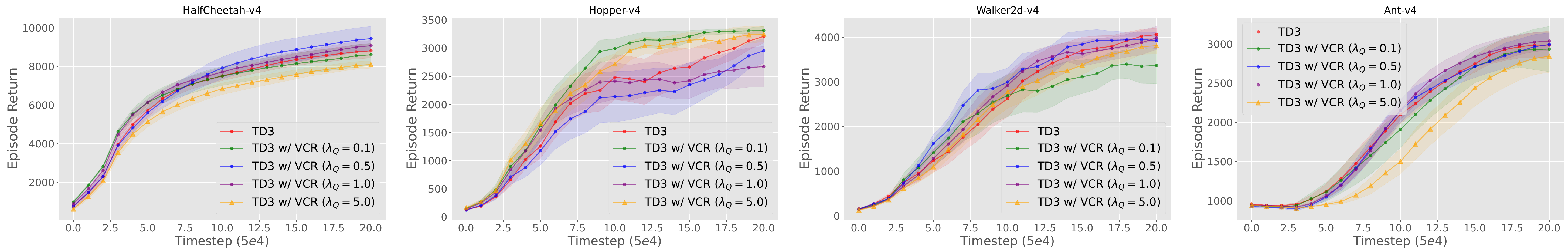}
}\\
\subfigure[Policy Churn Reduction (PCR)]{
\includegraphics[width=1.0\textwidth]{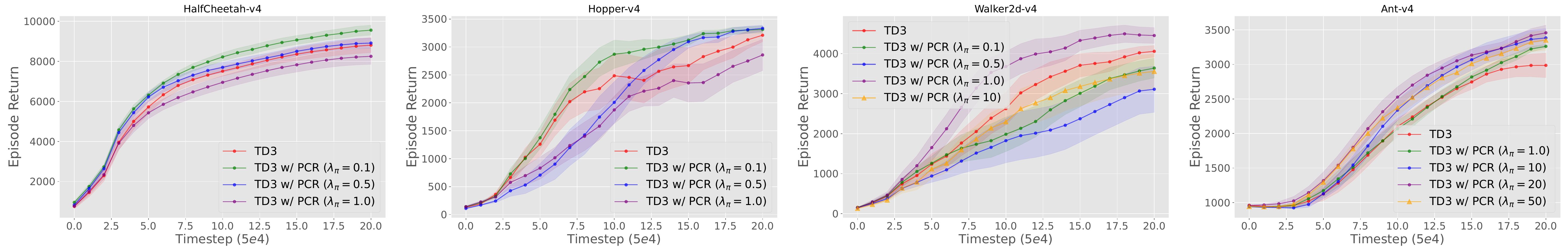}
}
\subfigure[Double Churn Reduction (DCR = VCR + PCR)]{
\includegraphics[width=1.0\textwidth]{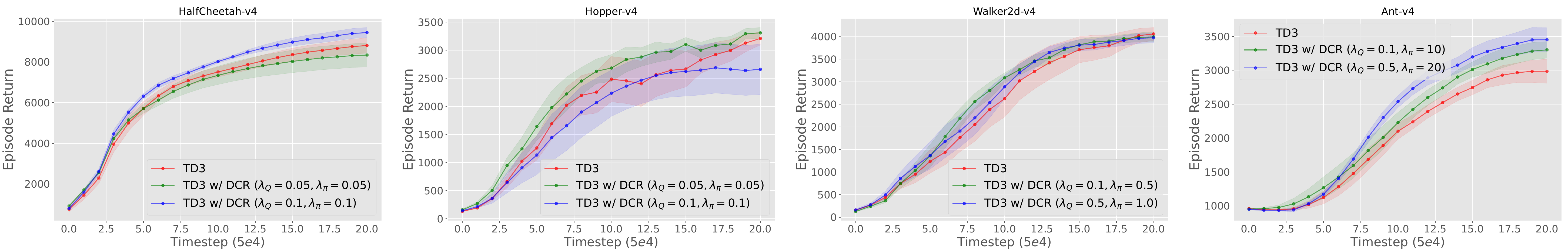}
}
\end{center}
\vspace{-0.3cm}
\caption{Hyperparameter choices for the value and policy churn reduction regularization for TD3 in MuJoCo. Curves and shades denote means and standard errors across six random seeds.
}
\label{figure:mujoco_td3_hyperparam_ana}
\end{figure*}

\begin{figure*}[ht]
\begin{center}
\subfigure[Value Churn Reduction (VCR)]{
\includegraphics[width=1.0\textwidth]{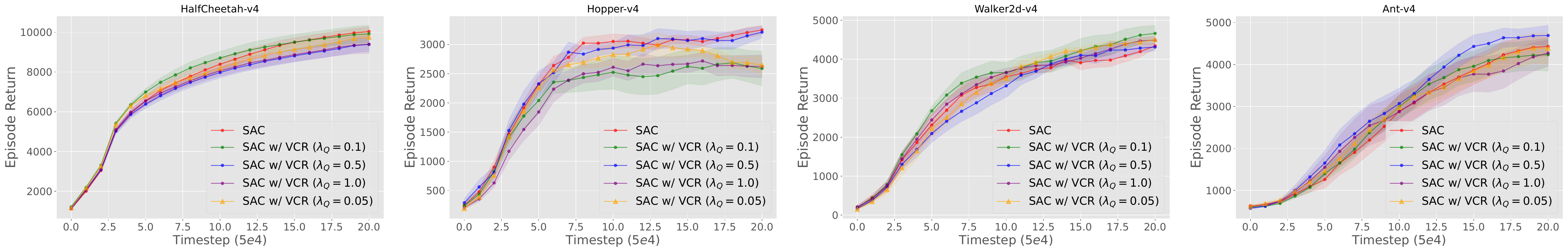}
}\\
\subfigure[Policy Churn Reduction (PCR)]{
\includegraphics[width=1.0\textwidth]{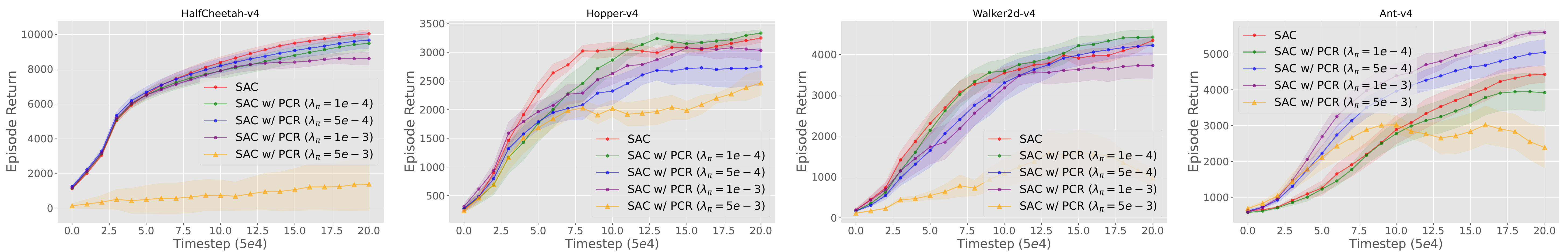}
}
\subfigure[Double Churn Reduction (DCR = VCR + PCR)]{
\includegraphics[width=1.0\textwidth]{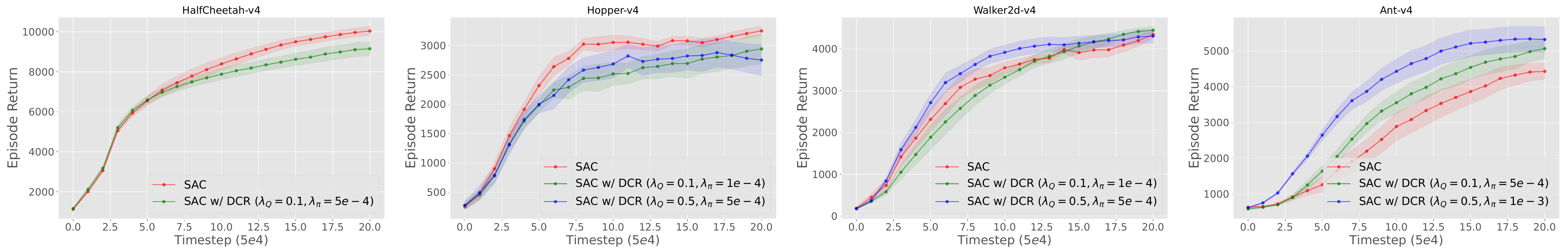}
}
\end{center}
\vspace{-0.3cm}
\caption{Hyperparameter choices for the value and policy churn reduction regularization for SAC in MuJoCo. Curves and shades denote means and standard errors across six random seeds.
}
\label{figure:mujoco_sac_hyperparam_ana}
\end{figure*}

\begin{figure*}[t]
\begin{center}
\includegraphics[width=1.0\textwidth]{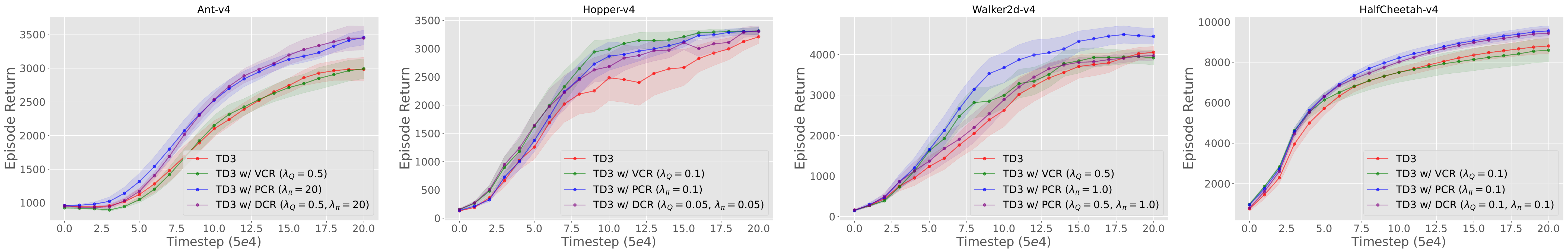}
\includegraphics[width=1.0\textwidth]{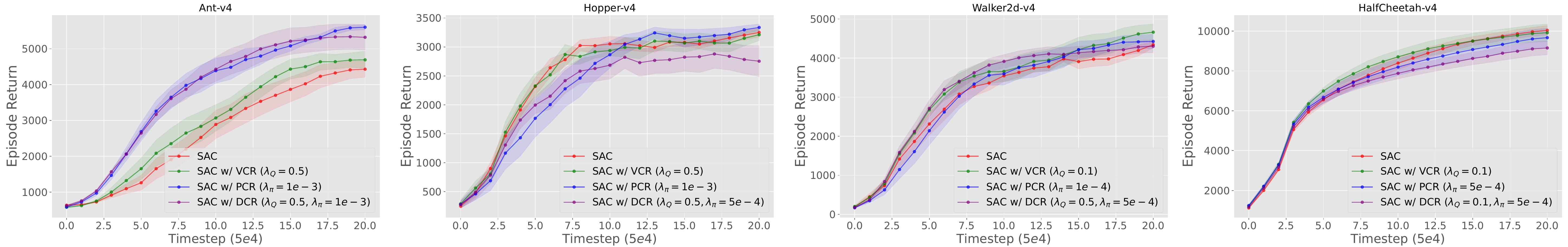}
\end{center}
\vspace{-0.2cm}
\caption{An overall comparison among using value churn reduction, using policy churn reduction, and using them both
for TD3 and SAC.}
\vspace{-0.1cm}
\label{figure:mujoco_eval_full_appendix}
\end{figure*}

\begin{figure*}[h]
\begin{center}
\includegraphics[width=1.0\textwidth]{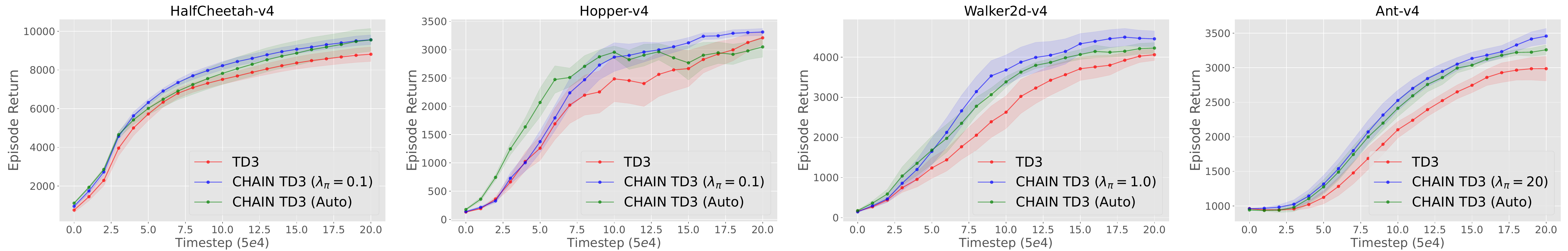}
\end{center}
\vspace{-0.4cm}
\caption{The results for auto-adjustment of $\lambda_{\pi}$ for CHAIN TD3 in four MuJoCo tasks, with means and standard errors across six seeds. The target relative loss scale $\beta$ is set to 5e-5. CHAIN (Auto) achieves comparable performance with manually selected coefficients and improves TD3.
}
\vspace{-0.2cm}
\label{figure:mujoco_td3_pcra_auto_eval}
\end{figure*}

\begin{figure*}[ht]
\vspace{-0.1cm}
\begin{center}
\subfigure[Final scores in Ant]{
\includegraphics[width=1.0\textwidth]{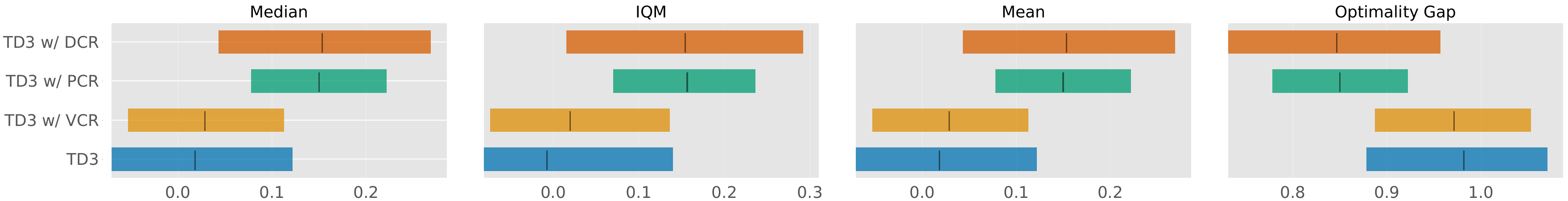}
}\\
\vspace{-0.2cm}
\subfigure[AUC scores in Ant]{
\includegraphics[width=1.0\textwidth]{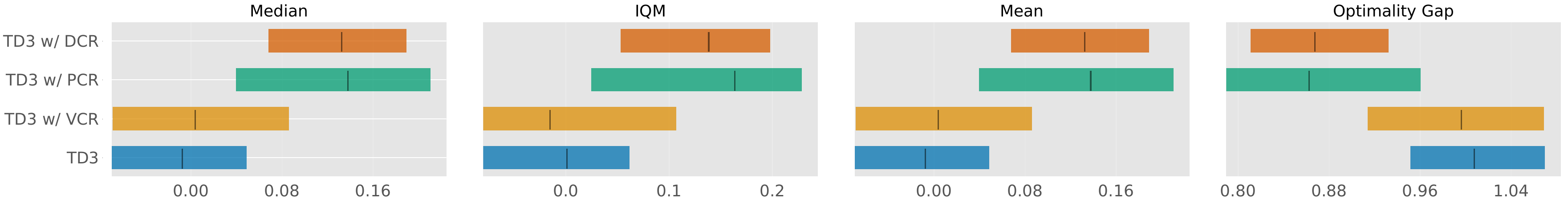}
}\\
\vspace{-0.2cm}
\subfigure[Final scores over all MuJoCo tasks]{
\includegraphics[width=1.0\textwidth]{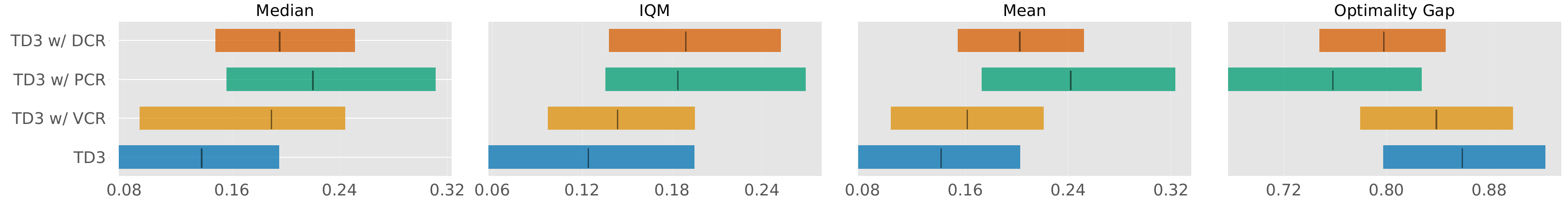}
}\\
\vspace{-0.2cm}
\subfigure[AUC scores over all MuJoCo tasks]{
\includegraphics[width=1.0\textwidth]{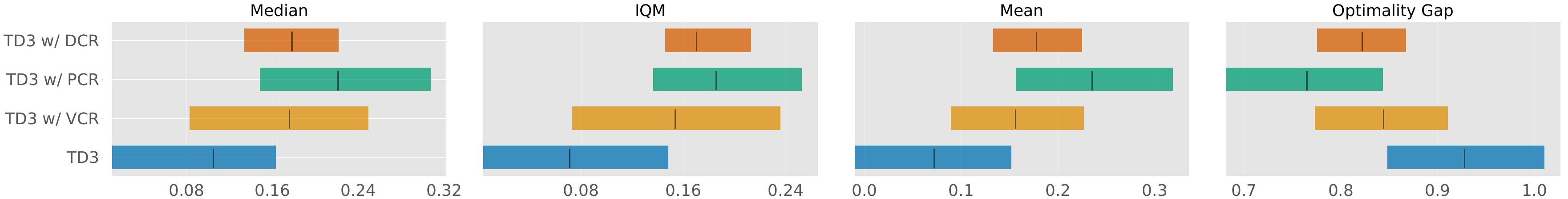}
}
\end{center}
\vspace{-0.3cm}
\caption{\textit{Reliable}~\citep{agarwal2021deep} metrics for TD3 with different churn reduction options.}
\vspace{-0.2cm}
\label{figure:mujoco_td3_reliable_agg_metrics}
\end{figure*}

\begin{figure*}[ht]
\vspace{-0.2cm}
\begin{center}
\subfigure[Final scores in Ant]{
\includegraphics[width=1.0\textwidth]{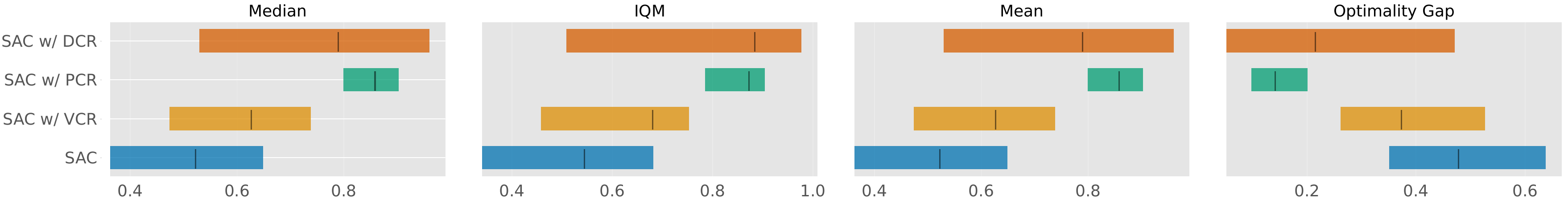}
}\\
\vspace{-0.2cm}
\subfigure[AUC scores in Ant]{
\includegraphics[width=1.0\textwidth]{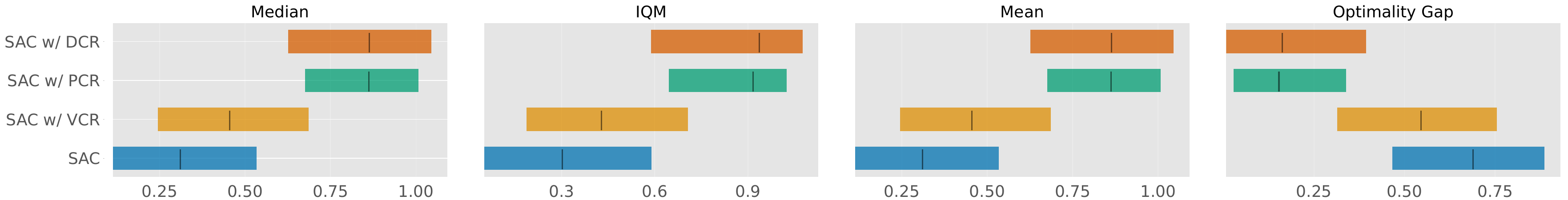}
}\\
\vspace{-0.2cm}
\subfigure[Final scores over all MuJoCo tasks]{
\includegraphics[width=1.0\textwidth]{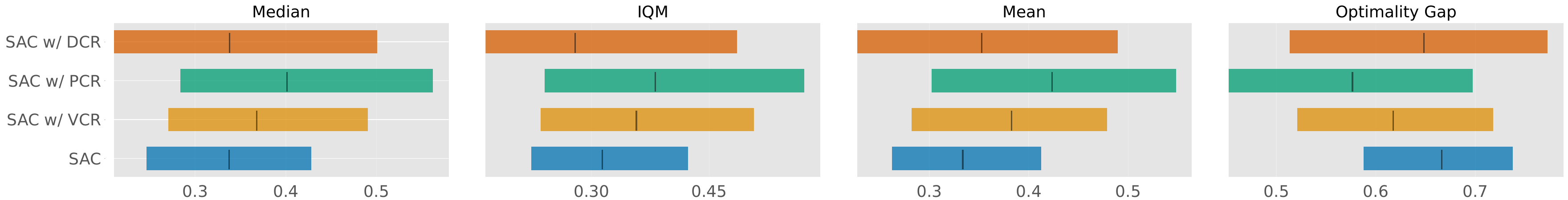}
}\\
\vspace{-0.2cm}
\subfigure[AUC scores over all MuJoCo tasks]{
\includegraphics[width=1.0\textwidth]{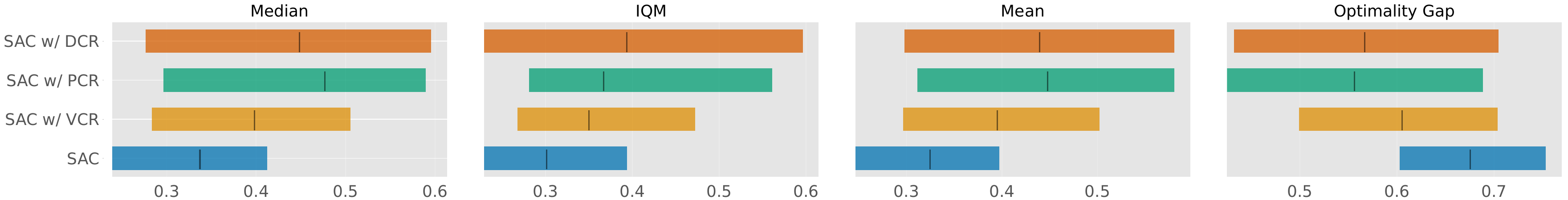}
}
\end{center}
\vspace{-0.3cm}
\caption{\textit{Reliable}~\citep{agarwal2021deep} metrics for SAC with different churn reduction options.
}
\vspace{-0.5cm}
\label{figure:mujoco_sac_reliable_agg_metrics}
\end{figure*}

We refer the readers to the following figures and tables for concrete additional results:
\begin{itemize}
    \item Figure~\ref{figure:mujoco_td3_churn_reduction} and Figure~\ref{figure:mujoco_sac_churn_reduction} show how the value churn reduction (VCR) and policy churn reduction (PCR) take effect during the learning process of TD3 and SAC.
    The four metrics defined in Table~\ref{table:churn_metrics} are used.
    
    \item Figure~\ref{figure:mujoco_td3_hyperparam_ana} and Figure~\ref{figure:mujoco_sac_hyperparam_ana} show the results of different choices of $\lambda_{Q}$ and $\lambda_{\pi}$ when VCR, PCR, or VCR+PCR (i.e., denoted by DCR) in CHAIN TD3 and CHAIN SAC.

    \item Figure~\ref{figure:mujoco_eval_full_appendix} shows an overall comparison among VCR, PCR, DCR when using either of them in CHAIN TD3 and CHAIN SAC. 
    We also provide the comparison with \textit{Reliable} Metrics\footnote{\url{https://github.com/google-research/rliable}}~\citep{agarwal2021deep} in Figure~\ref{figure:mujoco_td3_reliable_agg_metrics} and Figure~\ref{figure:mujoco_sac_reliable_agg_metrics}.

    \item 
    Figure~\ref{figure:mujoco_td3_pcra_auto_eval} shows the evaluation of the automatic adjustment method of $\lambda_{\pi}$ for CHAIN TD3 in four MuJoCo tasks.
    
\end{itemize}

\paragraph{The Effect of CHAIN-VCR and CHAIN-PCR in Reducing Churn}

For TD3 and SAC, the amount of value churn (\textcolor{blue}{$\hat{\mathcal{C}}_{|Q_{sa}|}$}) and policy churn (\textcolor{blue}{$\hat{\mathcal{C}}_{\pi}$}) increases as the update number $i$ increases.
This is expected as the chain effect indicates the accumulation of churn and deviation. 
Besides, the churn saturates after a sufficient number of updates on the networks. We hypothesize the target network helps cap the amount of churn.

Moreover, we observe a positive policy value deviation of action churn (\textcolor{blue}{$\hat{\mathcal{D}}^{\pi}_{Q}$}), which matches our discussion on \textit{implicit optimization} effect of the policy churn in Appendix~\ref{app:derivations}; in contrast, the value churn (\textcolor{blue}{$\hat{\mathcal{C}}_{Q_{sa}}$}) fluctuates above and below 0.

According to the relative scale, TD3 and SAC show less value and policy churn than DoubleDQN. This is due to the structure independence of actor and critic networks. Another possible reason is that MuJoCo locomotion may have smoother underlying problem dynamics than the MinAtar.

\paragraph{Discussion on the Effect of CHAIN-VCR, -PCR, -DCR in Improving Episode Return}

In addition to the two variants of CHAIN introduced in the main body of this paper,
we introduce the third one, i.e., Double Churn Reduction (DCR), which corresponds to applying VCR and PCR at the same time.
The related results are reported in Figure~\ref{figure:mujoco_td3_hyperparam_ana},~\ref{figure:mujoco_sac_hyperparam_ana},~\ref{figure:mujoco_eval_full_appendix}.

For CHAIN-VCR and CHAIN-PCR,
we found that CHAIN-PCR often improves the learning performance, especially for Ant-v4. In contrast, CHAIN-VCR improves slightly.
We hypothesize that this is because policy interacts with the environment directly and the target critic-network also helps to cap the value churn.
Between TD3 and SAC, CHAIN-PCR works better for TD3 rather than SAC. We hypothesize that
the variation (regarding the scale and range) is higher in optimizing the Maximum-Entropy objective and KL-based PCR term together for SAC than in optimizing the $Q$ objective and L2-based PCR term for TD3. Another hypothesis is that the Maximum-Entropy nature of SAC prefers the encouragement of more stochasticity in policy.

For CHAIN-DCR, we found that it is not easy to gain an immediate additive improvement regarding episode return when using the same hyperparameter choices $\lambda_{Q},\lambda_{\pi}$ from both sides.
We suggest that this reflects the intricate nature of the chain effect.
As mentioned in \Cref{sec:conclusion}, this points out the limitation of this work in aspects like the in-depth theoretical analysis of the long-term chain effect,
the lack of automatic coefficient adjustment, and the finer-grained understanding of the positive and negative effects of churn.
We leave these directions for future work.

\clearpage

\paragraph{The Effect of Automatic Adjustment of $\lambda_{\pi}$}

Figure~\ref{figure:mujoco_td3_pcra_auto_eval} shows the evaluation of the automatic adjustment method of $\lambda_{\pi}$ for CHAIN TD3 in four MuJoCo tasks.
We can observe that CHAIN (Auto) achieves comparable performance with manually selected coefficients and improves TD3, similarly to the conclusions we found in Figure~\ref{figure:minatar_auto_eval}  and~\ref{figure:mujoco_and_dmc_auto_eval}.

\clearpage

\subsection{More Results of CHAIN for IQL with Sequential Training}
\label{app:more_for_iql}

As mentioned in Section~\ref{subsec:exp_deep_ac}, the policy network of IQL has no impact on the training of the value networks, since the value networks (i.e., $Q$ and $V$) are trained purely based on in-sample data without accessing $a^{\prime} = \pi_{\phi}(s^{\prime})$. Thus, although the policy and value networks of IQL still have churns, the chain effect of churn does not apply in this case.

The default implementation of IQL follows the fashion of iterative training between the policy network and the value network(s).
Since the training of the value networks of IQL is independent of the policy, one natural idea is to first fully train the value networks for a sufficient budget and then train the policy value with the well-trained and frozen value networks.
We call this \textit{actor-trained-against-final-frozen-critic} fashion as \textbf{sequential} training of IQL.

We slightly modified the training process of the CORL implementation of IQL to realize the sequential training of IQL:
(1) First train the value network and Q network for 1M steps;
(2) Then train the policy network for 1M steps with the value network and Q network frozen;
(3) We do not modify any other implementation detail and use the same hyperparameters;
(4) We check the learning curves of the policy network and the final scores.
We call this variation \textbf{IQL (sequential)}. The total number of gradient step is the same as the default IQL implementation where the critic and actor are trained iteratively.

We report the final scores of IQL (sequential) with means and standard errors over 12 seeds in Table~\ref{table:chain_offline_iql_plus}.
The results show that IQL (sequential) performs worse than IQL in 5 of 6 tasks.
The difference between IQL (sequential) and IQL can be fully attributed to the difference in the training dynamics, mainly on the policy network.
This also means exposing the actor network to the training process of the $Q$ network is helpful compared to training the actor based on a sufficiently trained and frozen $Q$ network.
To provide some possible explanation, we guess that this is because the final $Q$ network is a product of accumulated value churns, and the difference in the training dynamics of the policy network of IQL results in a further difference in the policy output on out-of-sample states.
However, it is somewhat tricky to explain the difference between IQL (sequential) and IQL. The dynamics is beyond the scope of the chain effect of churn mechanism studied in our work.

\begin{table}[h]
  \caption{Results for IQL, IQL (sequential) and CHAIN IQL in Antmaze.}
  \vspace{0.0cm}
  \centering
  \scalebox{0.9}{
  \begin{tabular}{c|c|c|c|c}
    \toprule
    Task & IQL & IQL (sequential) & CHAIN IQL (PCR) & CHAIN IQL (VCR) \\
    \midrule
    AM-umaze-v2 & 77.00 $\pm$ 5.52 & 60.00 $\pm$ 3.91 & \textbf{84.44 $\pm$ 3.19} & 83.33 $\pm$ 2.72\\
    AM-umaze-diverse-v2 & 54.25 $\pm$ 5.54 & 55.00 $\pm$ 5.46 & 62.50 $\pm$ 3.75 & \textbf{71.67 $\pm$ 7.23} \\
    AM-medium-play-v2 & 65.75 $\pm$ 11.71 & 52.50 $\pm$ 3.36 & \textbf{72.50 $\pm$ 2.92} & 70.00 $\pm$ 3.33 \\
    AM-medium-diverse-v2 & 73.75 $\pm$ 5.45 & 53.33 $\pm$ 5.93 & \textbf{76.67 $\pm$ 4.51} & 66.67 $\pm$ 3.79 \\
    AM-large-play-v2 & 42.00 $\pm$ 4.53 & 17.5  $\pm$  4.10 & \textbf{50.00 $\pm$ 4.56} & 43.33 $\pm$ 4.14 \\
    AM-large-diverse-v2 & 30.25 $\pm$ 3.63 & 5.83 $\pm$ 2.19 & 26.67 $\pm$ 3.96 & \textbf{31.67 $\pm$ 2.31} \\
    \bottomrule
  \end{tabular}
  }
\label{table:chain_offline_iql_plus}
\end{table}

\subsection{More Results for Scaling PPO with CHAIN}
\label{app:more_for_ppo_scaling}

We take PPO and MuJoCo tasks as the exemplary setting and widen both the policy and value networks by a scale-up ratio within $\{2, 4, 8, 16\}$. Note that the default network architecture (i.e., when the scale-up ratio equals one) for both the policy and value networks is a two-layer MLP with 256 neurons for each layer, followed by an output layer.

\begin{figure*}
\begin{center}
\includegraphics[width=1.0\textwidth]{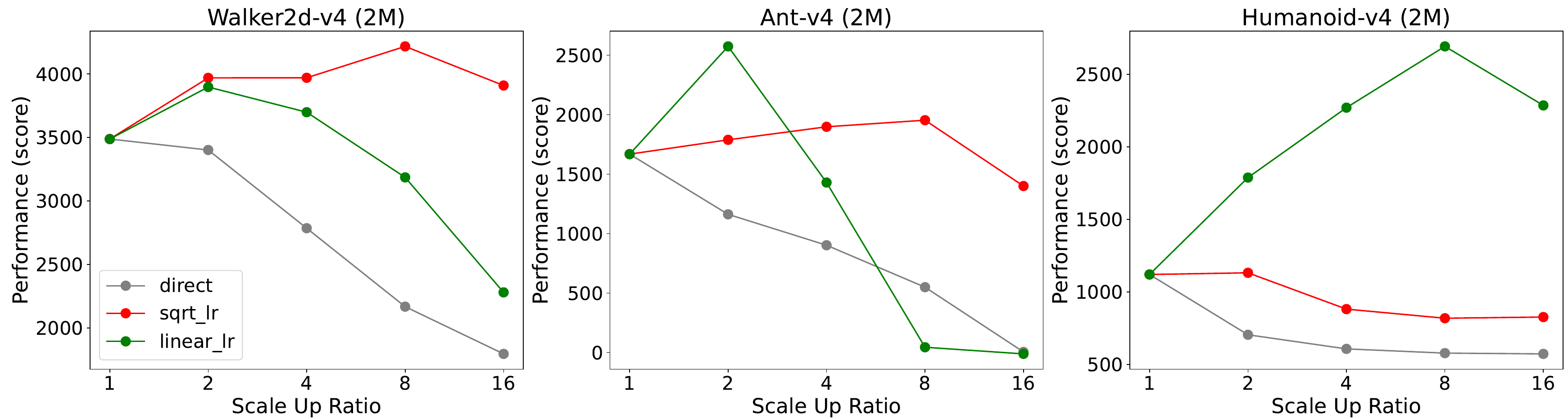}
\includegraphics[width=0.65\textwidth]{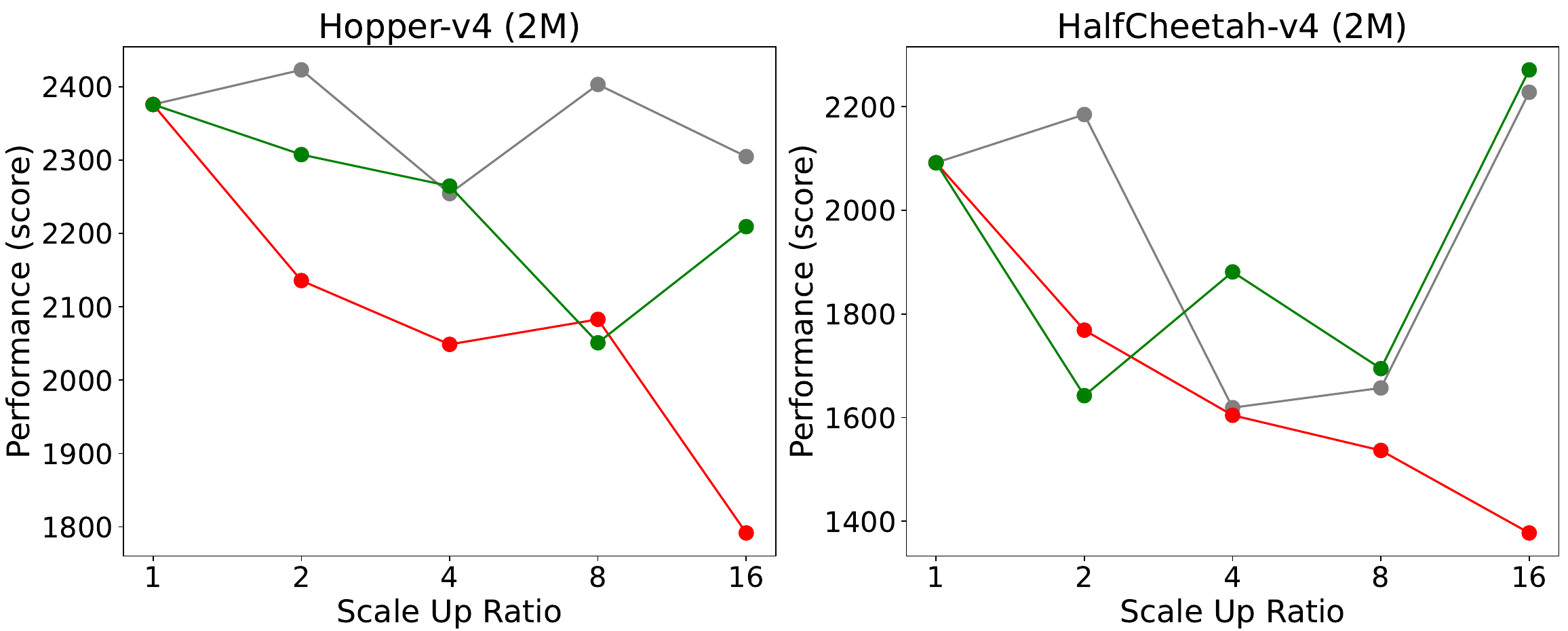}
\end{center}
\vspace{-0.2cm}
\caption{The results for PPO scaling by \textit{widening} with different learning rate settings.
`direct' means using the default learning rate, `linear' means using a decreased learning rate as \texttt{lr / scale-up ratio}, and `sqrt' means using \texttt{lr / sqrt(scale-up ratio)}.}
\vspace{-0.4cm}
\label{figure:ppo_scaling_lr}
\end{figure*}

Inspired by the prior study~\citep{Ob24Pruned},
in addition to directly scaling PPO up (`direct'),
we use two variants that use different learning rate settings for comparison:
(1) `linear' means using a decreased learning rate as \texttt{lr / scale-up ratio}, and (2) `sqrt' means using \texttt{lr / sqrt(scale-up ratio)}.
The results of scaling PPO with different learning rate settings are shown in Figure~\ref{figure:ppo_scaling_lr} by different colors.

As expected, we observed that the performance of PPO degrades as the increase of the scale-up ratio severely.
We found using a decreased learning rate with `linear' or `sqrt' alleviates the degradation of PPO scaling to some degree.

The evaluation results of the effect of CHAIN in this scaling setting are shown in Figure~\ref{figure:ppo_scaling_pcr}, Figure~\ref{ppo_scaling_pcr_full} and Table~\ref{table:ppo_scaling_with_chain} as discussed in the main body of this paper.

\begin{figure*}[h]
\begin{center}
\subfigure[PPO Scaling by \textit{Widening}]{\includegraphics[width=0.9\textwidth]{figs/mujoco_scale_ppo_pcr_full_comparison_240819.pdf}
\label{subfig:ppo_scale_wide}
}\\
\vspace{-0.2cm}
\subfigure[PPO Scaling by \textit{Deepening}]{
\includegraphics[width=0.9\textwidth]{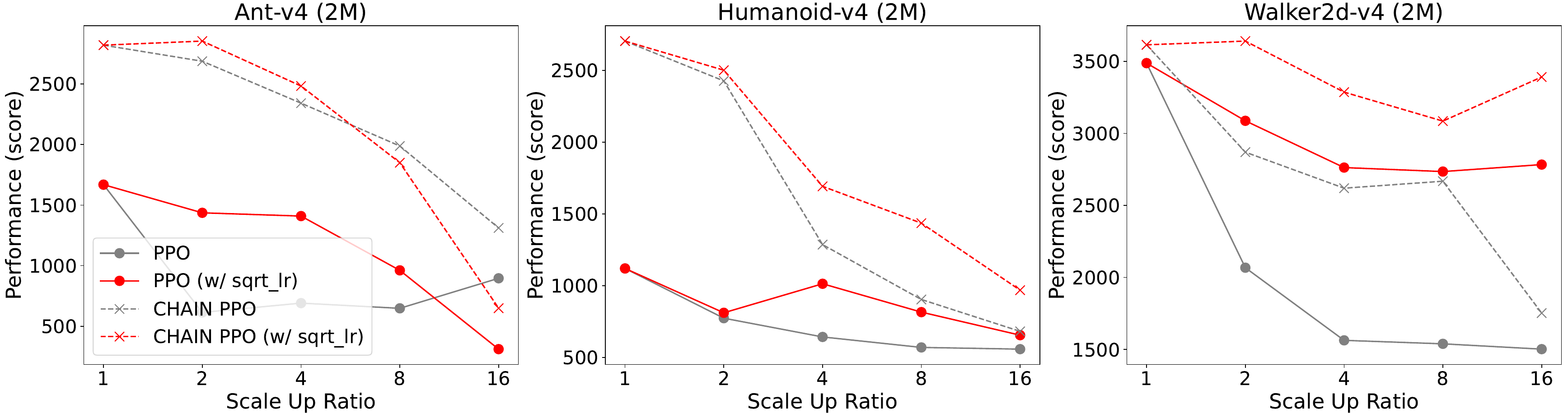}
\label{subfig:ppo_scale_deep}
}
\end{center}
\vspace{-0.4cm}
\caption{The results in terms of episode return for scaling PPO with CHAIN. CHAIN helps to scale almost across all the configurations.}
\vspace{-0.2cm}
\label{figure:ppo_scaling_pcr_full}
\end{figure*}

\subsection{More Results for CHAIN v.s. Slowing Down Learning}
\label{app:more_for_slower_lraning}

Since churn accompanies each time mini-batch training performed for the networks, a natural question is: whether slowing down learning can alleviate the issue of churn.
In principle, using smaller learning rates or target network replacement rates should lead to less churn. This is because churn is positively related to the parameter update amount (as shown by the NTK expressions in Eq.~\ref{eq:churn_caused_by_parameter_update}) and a slower target network also slows the churn that occurs instantly in each training more when computing the target value to fit for the critic-network.

Empirically, we ran DDQN in Asterix and Freeway and TD3 in Walker2d and Ant, with different learning rates and target network replacement rates. The results are summarized in Table~\ref{table:diff_lr_and_trr}.
We can observe that either reducing learning rate or target network replacement rate often leads to worse performance, especially for TD3. To some extent, this also matches the common knowledge in the RL community.

\begin{table}[h]
\caption{Different learning rates (\textit{lr}) and target network replacement rates (\textit{trr}), e.g., "/ 2" means "divided by 2". Mean final episode returns over six seeds are reported.
  }
  \vspace{-0.1cm}
  \centering
  \scalebox{0.9}{
  \begin{tabular}{c|cc|c|cc}
    \toprule
    Alg. - Task & Walker2d & Ant & Alg. - Task & Asterix & Freeway \\
    \midrule
    TD3 &  \textbf{4059.45} & 3069.14 & DDQN & 15.05 & 49.21 \\
    \midrule
    TD3 (\textit{lr} / 2) & 2774.87 & 2408.75 & DDQN (\textit{lr} / 2) & \textbf{19.26} & \textbf{55.83}  \\
    TD3 (\textit{lr} / 10) & 1314.12 & 1023.78 & DDQN (\textit{lr} / 10) & 16.20  & 49.20 \\
    \midrule
    TD3 (\textit{trr} / 5) & 3170.18 & \textbf{3375.33} & DDQN (\textit{trr} / 5) & 11.36 & 50.76 \\
    TD3 (\textit{trr} * 5) & 4057.33 & 2743.25 & DDQN (\textit{trr} * 5) & 18.26 & 54.33 \\
    \bottomrule
  \end{tabular}
  }
\label{table:diff_lr_and_trr}
\vspace{-0.1cm}
\end{table}

This indicates that the issue of churn cannot be addressed by reducing learning rate or target network replacement rate (which usually slows down the learning process). Churn is a “by-product” of the training of DRL agents and should be addressed separately.

\clearpage
\newpage
\section*{NeurIPS Paper Checklist}

\begin{enumerate}

\item {\bf Claims}
    \item[] Question: Do the main claims made in the abstract and introduction accurately reflect the paper's contributions and scope?
    \item[] Answer: \answerYes{} %
    \item[] Justification: The main claims in the abstract and introduction summarize our contributions and are supported by our formal analysis in~\Cref{sec:churn} and our experiment results in~\Cref{sec:exps}.

\item {\bf Limitations}
    \item[] Question: Does the paper discuss the limitations of the work performed by the authors?
    \item[] Answer: \answerYes{} %
    \item[] Justification: We discuss the limitations in Appendix~\ref{app:limitations}.

\item {\bf Theory Assumptions and Proofs}
    \item[] Question: For each theoretical result, does the paper provide the full set of assumptions and a complete (and correct) proof?
    \item[] Answer: \answerNA{} %
    \item[] Justification: This paper does not include theoretical results.

    \item {\bf Experimental Result Reproducibility}
    \item[] Question: Does the paper fully disclose all the information needed to reproduce the main experimental results of the paper to the extent that it affects the main claims and/or conclusions of the paper (regardless of whether the code and data are provided or not)?
    \item[] Answer: \answerYes{} %
    \item[] Justification: We provide sufficient experiment and implementation details in Section~\ref{sec:exps} and Appendix~\ref{app:exp_details}.

\item {\bf Open access to data and code}
    \item[] Question: Does the paper provide open access to the data and code, with sufficient instructions to faithfully reproduce the main experimental results, as described in supplemental material?
    \item[] Answer: \answerYes{} %
    \item[] Justification: Our code can be found at \url{https://github.com/bluecontra/CHAIN}.

\item {\bf Experimental Setting/Details}
    \item[] Question: Does the paper specify all the training and test details (e.g., data splits, hyperparameters, how they were chosen, type of optimizer, etc.) necessary to understand the results?
    \item[] Answer: \answerYes{} %
    \item[] Justification: We provide necessary experimental details in Section~\ref{sec:exps} and Appendix~\ref{app:exp_details}.

\item {\bf Experiment Statistical Significance}
    \item[] Question: Does the paper report error bars suitably and correctly defined or other appropriate information about the statistical significance of the experiments?
    \item[] Answer: \answerYes{} %
    \item[] Justification: The results in our paper are accompanied by error bars.

\item {\bf Experiments Compute Resources}
    \item[] Question: For each experiment, does the paper provide sufficient information on the computer resources (type of compute workers, memory, time of execution) needed to reproduce the experiments?
    \item[] Answer: \answerYes{} %
    \item[] Justification: We provide the details of compute resources in Appendix~\ref{app:exp_details}.
    
\item {\bf Code Of Ethics}
    \item[] Question: Does the research conducted in the paper conform, in every respect, with the NeurIPS Code of Ethics \url{https://neurips.cc/public/EthicsGuidelines}?
    \item[] Answer: \answerYes{} %
    \item[] Justification: This work conforms with the NeurIPS Code of Ethics.

\item {\bf Broader Impacts}
    \item[] Question: Does the paper discuss both potential positive societal impacts and negative societal impacts of the work performed?
    \item[] Answer: \answerNA{} %
    \item[] Justification: This paper presents work whose goal is to advance the field of Deep Reinforcement Learning (DRL). No specific real-world application is concerned. Our study on the churn phenomenon may help us better understand and control the behaviours of DRL agents, which can have a positive impact in general on the applicability and safety of DRL techniques.
    
\item {\bf Safeguards}
    \item[] Question: Does the paper describe safeguards that have been put in place for responsible release of data or models that have a high risk for misuse (e.g., pretrained language models, image generators, or scraped datasets)?
    \item[] Answer: \answerNA{} %
    \item[] Justification: The paper poses no such risks.

\item {\bf Licenses for existing assets}
    \item[] Question: Are the creators or original owners of assets (e.g., code, data, models), used in the paper, properly credited and are the license and terms of use explicitly mentioned and properly respected?
    \item[] Answer: \answerYes{} %
    \item[] Justification: All the environments, data and codes of baseline methods used in this paper are publicly available on Github. We cited the original papers and provided the URLs to the assets.

\item {\bf New Assets}
    \item[] Question: Are new assets introduced in the paper well documented and is the documentation provided alongside the assets?
    \item[] Answer: \answerYes{} %
    \item[] Justification: We provide the necessary details for implementing our proposed method in this paper. We will provide a README document alongside the code to release after the review process.

\item {\bf Crowdsourcing and Research with Human Subjects}
    \item[] Question: For crowdsourcing experiments and research with human subjects, does the paper include the full text of instructions given to participants and screenshots, if applicable, as well as details about compensation (if any)? 
    \item[] Answer: \answerNA{} %
    \item[] Justification: The paper does not involve crowdsourcing nor research with human subjects.

\item {\bf Institutional Review Board (IRB) Approvals or Equivalent for Research with Human Subjects}
    \item[] Question: Does the paper describe potential risks incurred by study participants, whether such risks were disclosed to the subjects, and whether Institutional Review Board (IRB) approvals (or an equivalent approval/review based on the requirements of your country or institution) were obtained?
    \item[] Answer: \answerNA{} %
    \item[] Justification: The paper does not involve crowdsourcing nor research with human subjects.

\end{enumerate}

\end{document}